%% file: main.tex
\documentclass[10pt,journal,compsoc]{IEEEtran}

\usepackage{graphicx}
\usepackage{amsmath,amssymb}
\usepackage{multirow}
\usepackage{bm}

\usepackage{ragged2e}
\usepackage{algorithm}
\usepackage{algpseudocode}
\usepackage{booktabs}
\usepackage{colortbl}
\usepackage[accsupp]{axessibility}
\usepackage[numbers,sort&compress]{natbib}
\usepackage[shortlabels,inline]{enumitem}

\usepackage{hyperref}
\hypersetup{
	final,
	colorlinks,
}
\usepackage{url}
\usepackage[capitalize]{cleveref}

\input{preamble}

\def\ourMethod{{UniView}}

\definecolor{light}{gray}{0.95}

\setlist[itemize]{noitemsep,leftmargin=*,topsep=0em}
\setlist[enumerate]{noitemsep,leftmargin=*,topsep=0em}

\crefname{section}{Sec.}{Secs.}
\Crefname{section}{Section}{Sections}
\crefname{table}{Tab.}{Tabs.}
\Crefname{table}{Table}{Tables}

\begin{document}

\title{\ourMethod: Large-Baseline View Synthesis via Video Diffusion Models}

\author{
Wangbo~Yu\textsuperscript{*},
Haiyang~Zhou\textsuperscript{*},
Chaoran~Feng\textsuperscript{*},
Xunyu~Zhou,
and~Li~Yuan\thanks{\textsuperscript{*}Wangbo Yu, Haiyang Zhou, and Chaoran Feng contributed equally to this work.
Wangbo Yu, Chaoran Feng, and Li Yuan are with Peking University. Haiyang Zhou is with Peking University and Rabbitpre AI\@. Xunyu Zhou is an intern at Peking University.}}

\markboth{}{}

\input{sections/0_abstract}

\maketitle

\IEEEdisplaynontitleabstractindextext

\IEEEpeerreviewmaketitle

\input{sections/1_introduction}
\input{sections/2_related}
\input{sections/3_method}
\input{sections/4_experiment}
\input{sections/5_conclusion}

{\small
\bibliographystyle{ieeenat_fullname}
\bibliography{main}
}

\end{document}

%% file: preamble.tex
\usepackage[dvipsnames]{xcolor}

\newcommand{\tablebodyfont}{\fontsize{8.94955}{10.85}\selectfont}

\makeatletter
\newcommand{\tablehighlight}{\rlap{\color{light}\vrule width\textwidth height\ht\@arstrutbox depth\dp\@arstrutbox}}
\makeatother

%% file: sections/0_abstract.tex
\IEEEtitleabstractindextext{
\begin{abstract}
\justifying
Novel view synthesis from casually captured monocular images and videos offers a practical route to immersive content creation without dedicated multi-view capture. However, the limited viewpoint coverage of such inputs leaves scene geometry insufficiently constrained, restricting the ability of reconstruction-based methods, such as neural radiance fields (NeRF) and 3D Gaussian Splatting (3DGS), to render views far from the observed viewpoints. Generative approaches leverage learned visual priors to synthesize unobserved content, but maintaining geometric consistency and precise camera control remains challenging under large viewpoint changes.
We present UniView, a unified framework for large-baseline novel view synthesis and scene generation from monocular images and videos. Central to our approach are an occlusion-aware point cloud renderer and a dual-branch conditioning mechanism. The renderer uses triple reprojection to identify and filter unreliable projections under large viewpoint changes. The conditioning mechanism integrates the resulting geometric guidance with source-view appearance within the video diffusion model, enabling precise camera control while preserving visual details and synthesizing unobserved content. To extend view synthesis to scene generation, we introduce a two-stage generation strategy that decouples viewpoint changes from temporal evolution. This strategy produces dense, synchronized multi-view videos that provide additional supervision for dynamic Gaussian Splatting, enabling more complete 4D generation from monocular observations.
Extensive evaluations across static and dynamic scenes demonstrate the effectiveness of UniView in novel view synthesis and scene generation, combining precise camera control with geometric consistency and visual fidelity.
Our code is available at \url{https://github.com/PKU-YuanGroup/UniView}.

\end{abstract}

}

%% file: sections/1_introduction.tex
\input{images_tex/teaser}

\section{Introduction}
\label{sec:intro}
Photorealistic novel view synthesis and scene generation in static 3D and dynamic 4D environments are fundamental to immersive content creation, with applications in VR/AR, robotics, gaming, and social media.
Reconstruction-based approaches, including neural radiance fields (NeRF)~\cite{mildenhall2020nerf,DynNeRF} and 3D Gaussian Splatting (3DGS)~\cite{kerbl20233dgs,4dgswu,wang2024shapeom}, achieve high rendering fidelity through per-scene optimization, but their reconstruction quality depends strongly on viewpoint coverage. With sparse observations, scene geometry and appearance remain poorly constrained in occluded or unseen regions, leading to rendering artifacts and geometric inconsistencies when synthesizing views far from the input viewpoints.

Recent advances in image and video diffusion models~\cite{rombach2022high,chen2023videocrafter1,blattmann2023svd,blattmann2023align,yang2024cogvideox,lin2024open} have introduced powerful generative priors for novel view synthesis, enabling plausible completion of unobserved regions from limited visual input. Building on these priors, camera-conditioned diffusion models synthesize novel views from a single image or a few input views~\cite{gao2024cat3d,zeronvs,seva,wang2023motionctrl,wu2024reconfusion,sun2024dimensionx}, as well as monocular videos~\cite{bai2025recammaster,gcd}. Despite their visual quality, conditioning on camera poses alone leaves the model to infer how scene content should project into the requested viewpoints. Without explicit geometric constraints, maintaining accurate camera control and cross-view consistency remains challenging, particularly under large viewpoint changes.

To provide direct geometric guidance, recent methods leverage visual geometry foundation models~\cite{wang2024dust3r,hu2024depthcrafter,dptv2,wang2025vggt,huang2025vipe,lan2025stream3r,wang2025moge} to recover 3D structure from monocular images and videos. A common paradigm is to estimate point clouds from monocular inputs and render them into geometric conditions for video diffusion models, thereby decoupling explicit view transformation from implicit content generation and enabling precise camera control~\cite{yu2024viewcrafter,ren2025gen3c,uni3c,you2024solver,muller2024multidiff,Ma2024See3D}.
While point clouds provide explicit geometric guidance for novel view synthesis, directly rendering them can fail to resolve visibility and occlusion relationships, particularly under large-baseline viewpoint changes. The resulting unreliable conditioning signals can compromise both training and inference, leading to geometric distortions and visual artifacts in the synthesized views.

To address these challenges, we present \ourMethod, a unified framework for large-baseline novel view synthesis and scene generation from monocular images and videos {\hypersetup{linkcolor=black,citecolor=black}(Fig.~\ref{fig:teaser})}.
The framework combines occlusion-aware geometric guidance with source-appearance conditioning for video diffusion. To obtain reliable geometric guidance under substantial viewpoint changes, we introduce an occlusion-aware point cloud renderer based on triple reprojection. By projecting the reconstructed geometry to the target views and back to the source views, we obtain source-view validity masks that filter the original point clouds before rendering the target views again. This procedure suppresses unreliable projections to provide geometric conditions for both training and inference. We adopt a dual-branch conditioning mechanism: the geometric-conditioning branch provides target-view spatial guidance, while the reference-injection branch retrieves source-view appearance information to preserve visual details. We train the model through a hybrid strategy that combines static multi-view datasets with large-scale dynamic monocular videos, supporting synthesis across both static and dynamic scenes.

Beyond novel view synthesis, we extend our framework to the generation of explicit 4D scene assets. We adopt a two-stage generation strategy that decouples viewpoint changes from temporal evolution to produce dense, temporally synchronized multi-view videos. These synthesized videos augment the original monocular observations with additional supervision for dynamic 3D Gaussian Splatting~\cite{4dgswu,lei2024mosca}, enabling the reconstruction of more complete 4D scenes.

Our contributions are summarized as follows:

\begin{itemize}
    \item We present {\ourMethod}, a unified framework for large-baseline view synthesis. We introduce a triple-reprojection-based point cloud renderer that uses source--target round-trip projections to identify unreliable regions and filter source points, reducing foreground--background tearing under large viewpoint changes.
    \item We adopt a dual-branch conditioning mechanism that combines target-view geometric guidance with source-view appearance information to promote cross-view consistency. To support training, we curate a data strategy that combines dynamic monocular videos with static multi-view resources, improving generalization and robustness across diverse scenes.
    \item We extend \ourMethod{} to explicit 4D scene generation by synthesizing dense, synchronized multi-view videos to supervise dynamic Gaussian Splatting. The resulting dynamic Gaussian scene representations support free-viewpoint rendering across time. Evaluations cover novel view synthesis and demonstrate improved rendering quality when generated views are used for 4D reconstruction.
\end{itemize}

This work extends TrajectoryCrafter~\cite{mark2025trajectorycrafter}, our preliminary conference version, in three aspects:
1) We substantially expand the range of novel view synthesis compared with TrajectoryCrafter~\cite{mark2025trajectorycrafter} by introducing a novel occlusion-aware point cloud rendering strategy that effectively resolves the occlusion ambiguity problem, which has long challenged previous point cloud--based methods~\cite{yu2024viewcrafter,ren2025gen3c,uni3c,you2024solver,muller2024multidiff}.
2) We scale our framework to a larger video diffusion backbone~\cite{vace,wan2025} to improve visual fidelity. Combined with the proposed rendering strategy, the upgraded framework supports high-quality view synthesis under extreme camera motions, including viewpoint changes of up to $360^\circ$.
3) We extend the framework beyond novel view synthesis to explicit 4D scene generation. A two-stage generation strategy produces dense, temporally synchronized multi-view videos from monocular inputs, providing additional supervision for dynamic Gaussian Splatting and enabling more complete 4D scene reconstruction.

%% file: images_tex/teaser.tex
\begin{figure*}[!t]
  \centering
  \includegraphics[width=1.\textwidth]{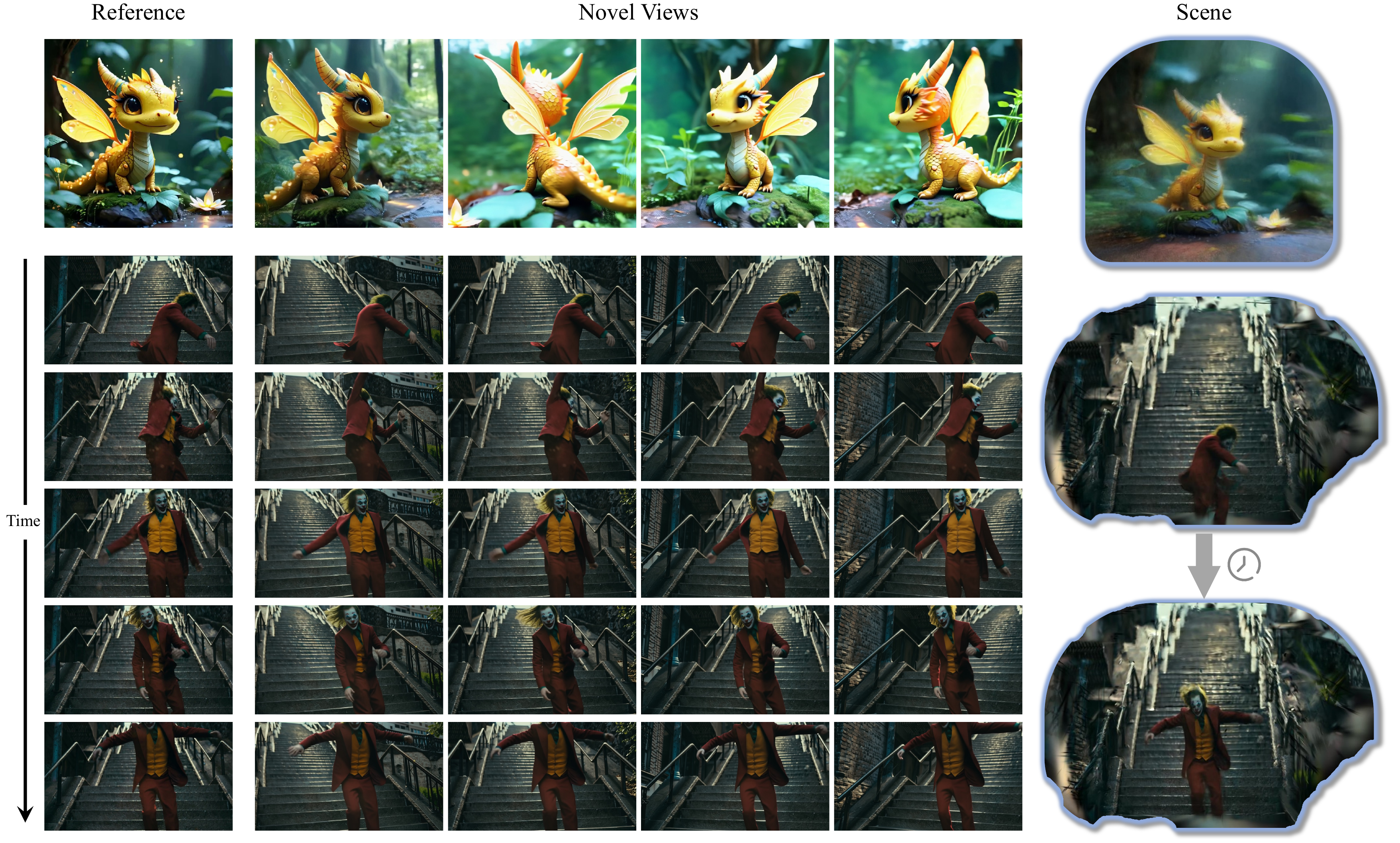}
  \caption{
  We present \ourMethod, a unified framework for controllable large-baseline novel view synthesis from monocular inputs.
  }
\label{fig:teaser}
\end{figure*}

%% file: sections/2_related.tex
\section{Related Work}
\label{sec:related}

\input{images_tex/pipeline}

\subsection{Reconstruction-based Novel View Synthesis}
The advent of neural representations like NeRF~\cite{mildenhall2020nerf} and 3DGS~\cite{kerbl20233dgs} has revolutionized static scene novel view synthesis~\cite{barron2021mip, barron2022mip,verbin2022ref,barron2023zip,hu2023Tri-MipRF,liu2024ripnerf,liang2025analytic,zhang2024pixelgs,yu2024mip,muller2022instant,fan2024instantsplat,yu2023nofa,yu2024evagaussians,feng2025ae,li2022nerfacc,chen2021mvsnerf,wang2023sparsenerf,zhu2023fsgs,lin2021barf,garbin2021fastnerf,chen2024mvsplat}.
For 4D novel view synthesis, existing approaches~\cite{hyperreel,fridovich2023kplane,cao2023hexplane,li2024spacetime,pumarola2021d,nerfplayer,yang2023gs4d} primarily focus on reconstructing 4D representations from synchronized multi-view videos, which are difficult to obtain for typical users.
Early efforts on 4D reconstruction from monocular videos relied on depth-based warping~\cite{yoon2020nvidia}, later refined with learned occlusion reasoning.
Subsequent studies~\cite{DynNeRF,sceneflow,nerfies,Tretschk_2021_ICCV,DynIBaR,lee2025fast} introduced neural representations for dynamic scene modeling, enabling improved reconstruction quality and temporal coherence.
Recent work~\cite{stearns2024marbles,wang2024shapeom,gao2024gaussianflow,lei2024mosca,xu20254dgt} further leverages the efficiency of 3DGS to synthesize novel views from monocular videos, often enhanced by auxiliary regularizations such as optical flow or depth for better spatiotemporal consistency.
However, recovering unobserved regions from monocular observations remains challenging, often leading to incomplete geometry and artifacts under large viewpoint changes.
Recent approaches~\cite{huang2025vivid4d,nazarczuk2025vidar,yang2026neoverse,zhu2026worldfrommotion} further combine 4D reconstruction with generative priors to synthesize views beyond the observed camera trajectory.

\subsection{Generative Novel View Synthesis}

Early works~\cite{wiles2020synsin,rombach2021geometryfree,rockwell2021pixelsynth,park2024bridging} integrate monocular depth estimation and image inpainting within a unified framework for novel view synthesis. However, these methods are generally constrained to category-specific domains such as object-centric or indoor scenes, and often produce visual artifacts due to their limited representation capacity.

The rapid advancement of diffusion models~\cite{ho2020denoising,song2021denoising,rombach2022high} has shown strong potential in generating novel views from monocular images and videos~\cite{zeronvs,huang2024roompainter,zhou2024holodreamer,yu2024hifi,liu2023zero,wang2023motionctrl}.
Building on this progress, recent works explore camera-pose-conditioned diffusion models for generative view synthesis.
Zero-1-to-3~\cite{liu2023zero} introduces a pose-conditioned diffusion model for object-level synthesis, which is later extended to general scenes by ZeroNVS~\cite{zeronvs}.
CAT3D~\cite{gao2024cat3d} leverages ray maps to enhance geometric consistency.
Subsequent studies further adopt video diffusion models~\cite{chen2023videocrafter1,blattmann2023svd,xing2023dynamicrafter,cogfun} for novel view synthesis, incorporating camera embeddings~\cite{wang2023motionctrl} or Plücker-coordinate encodings~\cite{bahmani2024vd3d,xu2024camco,bai2024syncammaster,he2024cameractrl,seva,liang2024wonderland,zhao2024genxd} to enable camera-controlled generation of static and dynamic scenes~\cite{gcd,wu2024cat4d,li2026cameraanything,huang2026spacetimepilot}.
These approaches learn the relationship between camera conditions and visual content, while maintaining geometric consistency under large viewpoint changes remains challenging.

With the rapid advancement of feed-forward visual-geometry estimation models~\cite{wang2024dust3r,hu2024depthcrafter,dptv2,wang2025vggt,huang2025vipe,lan2025stream3r,wang2025moge}, recent studies have increasingly incorporated explicit geometry into diffusion-based view synthesis.
Several works~\cite{gu2025das,xiao2024trajectory} integrate tracked points as 3D anchors to improve spatial consistency. Other works~\cite{wu2024reconfusion,wang2025videoscene,xie2026lavr} use features or renderings derived from reconstruction models~\cite{yu2021pixelnerf,chen2024mvsplat} to condition generation.
A number of approaches~\cite{yu2024viewcrafter,ren2025gen3c,uni3c,you2024solver,muller2024multidiff,Ma2024See3D,zhang2024recapture,yu2025wonderworld,wang2025vistadream,shriram2024realmdreamer,chung2023luciddreamer,liu2025free4d,tung2024megascenes,zhang2025spatialcrafter,lu2025see4d,lin2026vista4d,liu2026mocam,chen2026depthdirector,cai2026scaling4d} employ depth maps, geometric renderings, or cross-view correspondences to guide novel view synthesis.
Although these cues provide spatial guidance, errors in estimated geometry and ambiguous visibility can introduce misleading rendering conditions, particularly under large viewpoint changes.
Our work addresses this issue through occlusion-aware point cloud rendering that filters unreliable projections before conditioning the video diffusion model.

%% file: images_tex/pipeline.tex
\begin{figure*}[!t]
  \centering
  \includegraphics[width=1.\textwidth]{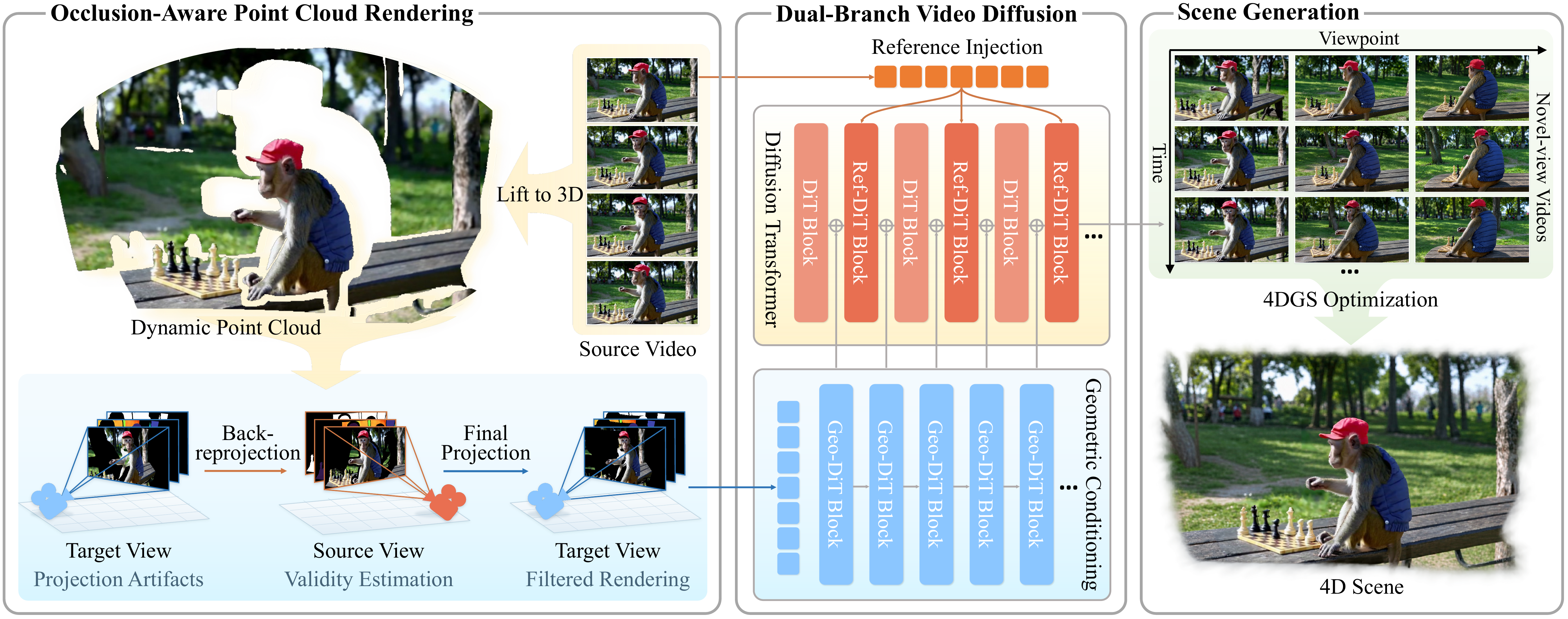}
  \caption{\textbf{Overview of {\ourMethod}.}
  Starting with a source video, whether casually captured or AI-generated, we first lift it into a dynamic point cloud via depth estimation. Users can then interactively render the point cloud with desired camera trajectories. The point cloud renders and the source video are jointly processed by our video diffusion model with dual-branch conditioning, yielding a high-fidelity video that precisely aligns with the specified trajectory and remains 4D consistent with the source video. For 4D reconstruction, we first synthesize views of the initial scene state and use them to anchor synchronized videos at fixed viewpoints. These videos and the source video jointly supervise the optimization of a dynamic Gaussian scene representation.
  }
\label{fig:pipeline1}
\end{figure*}

%% file: sections/3_method.tex
\section{Method}
\label{sec:method}

\input{images_tex/double_reprojection}

\subsection{Preliminary: Latent Video Flow Matching}
\label{subsec:preliminary}
Our generative backbone follows the latent flow-matching formulation of Wan~\cite{wan2025}. A pretrained video VAE encodes a clean video $\bm{x}\in\mathbb{R}^{n\times3\times h\times w}$ as $\bm{z}_0=\mathcal{E}(\bm{x})$, so that generation is modeled in the compressed spatiotemporal latent space.

Flow matching~\cite{lipman2023flow} learns a velocity field along paths connecting data and noise. For Gaussian noise $\bm{\epsilon}\sim\mathcal{N}(\bm{0},\bm{I})$ and noise level $t\in[0,1]$, we use the linear path
\begin{equation}
\bm{z}_t=(1-t)\bm{z}_0+t\bm{\epsilon},
\qquad
\frac{\mathrm{d}\bm{z}_t}{\mathrm{d}t}=\bm{\epsilon}-\bm{z}_0.
\label{eq:fm_path}
\end{equation}
A Diffusion Transformer (DiT)~\cite{dit} processes the patchified noisy latents and predicts their velocity $\bm{v}_\theta(\bm{z}_t,t\mid\bm{c})$, where $\bm{c}$ denotes the conditioning inputs. Text embeddings enter the backbone through cross-attention. With $t$ drawn from a training distribution $p(t)$, the velocity regression objective is
\begin{equation}
\mathcal{L}_{\mathrm{FM}}=
\mathbb{E}_{\bm{z}_0,\bm{c},t,\bm{\epsilon}}
\left[\left\|\bm{v}_\theta(\bm{z}_t,t\mid\bm{c})-(\bm{\epsilon}-\bm{z}_0)\right\|_2^2\right].
\label{eq:fm_loss}
\end{equation}

To generate a video, we initialize $\bm{z}_1$ with Gaussian noise and numerically integrate $\mathrm{d}\bm{z}_t/\mathrm{d}t=\bm{v}_\theta(\bm{z}_t,t\mid\bm{c})$ from $t=1$ to $t=0$. The resulting latent $\hat{\bm{z}}_0$ is decoded into the output video $\hat{\bm{x}}=\mathcal{D}(\hat{\bm{z}}_0)$.

\begin{figure}[!t]
  \centering
  \input{images_tex/normal}
\end{figure}

\subsection{Occlusion-aware Point Cloud Rendering}
\subsubsection{Point Cloud Reconstruction}
\label{subsec:pcd}
Given a source input $\bm{I}^{s} = \{\bm{I}_{i}^{s}\}_{i=1}^{n} \in \mathbb{R}^{n\times 3\times h\times w}$ (where the special case $n=1$ corresponds to a single-image input), we aim to explore its underlying scene with a desired camera trajectory.
To encourage precise camera trajectory control, we lift the source input into a point cloud and render novel views to serve as geometric conditioning for the diffusion model.
To implement this, we first utilize feed-forward geometry estimation models~\cite{zhang2024monst3r,lan2025stream3r,huang2025vipe} to estimate a sequence of depth maps $\bm{D}^{s} = \{\bm{D}_{i}^{s}\}_{i=1}^{n} \in \mathbb{R}^{n\times h\times w}$, camera poses $\bm{T}^{s} = \{\bm{T}^{s}_{i}\}_{i=1}^{n} \in \mathbb{R}^{n\times 4\times 4}$, and camera intrinsics $\bm{K} \in \mathbb{R}^{3\times 3}$.
Subsequently, we back-project the source input into a point cloud $\bm{P}^{s} = \{\bm{P}^{s}_{i}\}_{i=1}^{n}$:
\begin{equation}
    \bm{P}^{s} = \Phi^{-1}([\bm{I}^{s}, \bm{D}^{s}], \bm{T}^{s}, \bm{K}),
\label{eq:pcd}
\end{equation}
where $\Phi^{-1}$ denotes the inverse perspective projection.
We directly render the reconstructed point clouds from viewpoints specified by a target trajectory $\bm{T}^{r} = \{\bm{T}^{r}_{i}\}_{i=1}^{n} \in \mathbb{R}^{n\times 4\times 4}$. This initial projection yields rendered images $\bm{I}'$, depth maps $\bm{D}'$, and validity masks $\bm{M}'$:
\begin{equation}
    [\bm{I}', \bm{D}', \bm{M}'] = \Phi(\bm{P}^{s} \odot \bm{M}^{s}, \bm{T}^{r}, \bm{K}),
\label{eq:render}
\end{equation}
where $\Phi$ denotes the perspective projection.
$\bm{M}^{s} \in \mathbb{R}^{n\times h\times w}$ represents the validity mask of the source point cloud used to filter reliable input points, defaulting to an all-ones matrix.
The target-view validity masks $\bm{M}'$ distinguish valid rendered pixels from holes in $\bm{I}'$ caused by disocclusions and out-of-frame areas.
The masks  are binary validity masks: a value of $1$ denotes a valid pixel or point retained for rendering, whereas $0$ denotes an invalid element that is discarded.

The initial renders $\bm{I}'$ and their validity masks $\bm{M}'$ can directly condition the diffusion model, as explored in previous point-conditioned methods~\cite{mark2025trajectorycrafter,yu2024viewcrafter,Ma2024See3D,you2024solver}.
However, under large-baseline viewpoint changes, naive point cloud rendering can stretch foreground textures across background regions, producing foreground--background tearing (Fig.~\ref{fig:double_reprojection}, second row). Points on surfaces facing away from the target camera may also remain erroneously visible. These erroneous projections can mislead the video generation model. To obtain more reliable geometric conditioning, we use triple reprojection to disambiguate occlusions, with an additional surface-orientation check before the final rendering.

\subsubsection{Occlusion Disambiguation via Triple Reprojection}
\label{subsec:occlusion}

To suppress foreground--background tearing, we use triple reprojection: an initial source-to-target projection, a target-to-source reprojection to estimate source-view validity, and a final source-to-target projection after point filtering.

The first projection follows Eq.~\ref{eq:render}, producing target-view images $\bm{I}'$ (Fig.~\ref{fig:double_reprojection}, second row), depth maps $\bm{D}'$, and validity masks $\bm{M}'$. For the second projection, we back-project these images using their rendered depth maps to form intermediate point clouds $\bm{P}'$. We then render the points selected by $\bm{M}'$ along the source trajectory $\bm{T}^{s}$, obtaining back-reprojected images $\bm{I}''$ (Fig.~\ref{fig:double_reprojection}, third row), depth maps $\bm{D}''$, and validity masks $\bm{M}''$. Each mask is expressed in the image grid of its corresponding source frame, allowing it to filter the original source point cloud for that frame.

For a progressive transition from source to target viewpoints, we accumulate the back-reprojected masks $\bm{M}''$ across frames to preserve exclusions from earlier frames. The cumulative validity mask for frame $i$ is
\begingroup
\begin{equation}
\bm{M}^{\text{vis}}_{i} = \prod_{j=1}^{i} \bm{M}''_j, \qquad i \in \{1, \ldots, n\},
\end{equation}
\endgroup
where the elementwise product retains an image location only if it remains valid in all back-reprojection masks up to frame $i$. The accumulated masks identify which points in the original source point clouds are retained for the third projection back to the target viewpoints.

\begin{algorithm}[!t]
\caption{Occlusion-aware Point Cloud Rendering}\label{alg:occlusion}
\begin{algorithmic}[1]
\State {\textbf{Input:}}
\State \hspace{0.5cm}$ \text{Source video and depth } [\bm{I}^s, \bm{D}^s];$
\State \hspace{0.5cm}$ \text{Source valid mask } \bm{M}^s (\text{default: } \mathbf{1});$
\State \hspace{0.5cm}$ \text{Source and target trajectories } \bm{T}^s, \bm{T}^r, \text{ intrinsics } \bm{K};$
\State \hspace{0.5cm}$ \text{Normal-view alignment threshold } \alpha;$

\State \textbf{Reconstruct source point clouds}
\State \hspace{0.5cm}$ \bm{P}^s = \Phi^{-1}([\bm{I}^s, \bm{D}^s], \bm{T}^s, \bm{K}) $
\State \textbf{Projection 1: source $\rightarrow$ target}
\State \hspace{0.5cm}$ [\bm{I}', \bm{D}', \bm{M}'] = \Phi(\bm{P}^s \odot \bm{M}^s, \bm{T}^r, \bm{K}) $
\State \textbf{Projection 2: target $\rightarrow$ source}
\State \hspace{0.5cm}$ \bm{P}' = \Phi^{-1}([\bm{I}', \bm{D}'], \bm{T}^r, \bm{K}) $
\State \hspace{0.5cm}$ [\bm{I}'', \bm{D}'', \bm{M}''] = \Phi(\bm{P}'\odot \bm{M}', \bm{T}^s, \bm{K}) $
\State \textbf{Accumulate source-view validity masks}
\State \hspace{0.5cm}$ \bm{M}^{\text{vis}}_{i} = \prod_{j=1}^{i} \bm{M}''_j, \quad i \in \{1, \ldots, n\} $
\State \hspace{0.5cm}$ \bm{M}^{\text{vis}} = \{\bm{M}^{\text{vis}}_{i}\}_{i=1}^{n} $

\State \textbf{Normal-based visibility filtering}
\State \hspace{0.5cm}$ \mathbf{n} = \text{ComputeNormals}(\bm{P}^s) $
\State \hspace{0.5cm}$ \mathbf{v} = \text{ComputeViewVectors}(\bm{P}^s, \bm{T}^r) $
\State \hspace{0.5cm}$ \bm{M}^{\text{front}} = (\mathbf{n} \cdot \mathbf{v} > \cos(\alpha)) $

\State \textbf{Projection 3: source $\rightarrow$ target}
\State \hspace{0.5cm}$ [\bm{I}^{r}, \bm{D}^{r}, \bm{M}^{r}] = \Phi(\bm{P}^s \odot \bm{M}^{\text{vis}} \odot \bm{M}^{\text{front}}, \bm{T}^r, \bm{K}) $

\State {\textbf{Return:}}
\State \hspace{0.5cm}$ \text{Rendered Video and Depth } \bm{I}^{r}, \bm{D}^{r} $
\State \hspace{0.5cm}$ \text{Final Validity Mask } \bm{M}^{r} $
\end{algorithmic}
\end{algorithm}

Before the third projection, we additionally apply normal-based filtering to exclude points viewed from the back or at excessively oblique angles (Fig.~\ref{fig:normal}). For each source point $\mathbf{p}$, let $\mathbf{n}$ be its estimated unit normal oriented toward the source camera and $\mathbf{v}=(\mathbf{c}-\mathbf{p})/\|\mathbf{c}-\mathbf{p}\|$ the direction toward the target camera center $\mathbf{c}$. We define
\begingroup
\begin{equation}
    \bm{M}^{\text{front}}(\mathbf{p}) = \bigl(\mathbf{n}\cdot\mathbf{v}>\cos(\alpha)\bigr).
    \label{eq:front_face_mask}
\end{equation}
\endgroup
The mask equals $1$ when the condition holds and $0$ otherwise, where $\alpha\in[0,\pi/2]$ is the angular threshold. Combining this mask with $\bm{M}^{\text{vis}}$, the third projection yields the final images, depth maps, and validity masks:
\begin{equation}
    [\bm{I}^{r}, \bm{D}^{r}, \bm{M}^{r}] = \Phi\left(\bm{P}^{s} \odot \bm{M}^{\text{vis}} \odot \bm{M}^{\text{front}}, \bm{T}^{r}, \bm{K}\right).
    \label{eq:final_render}
\end{equation}
The resulting target-view renders are shown in Fig.~\ref{fig:double_reprojection} (fourth row). By filtering unreliable source points before the final projection, this procedure reduces foreground--background tearing and spurious back-face projections, providing more reliable geometric conditions for video generation. Regions without retained projections remain masked out for completion by the generative model. Algorithm~\ref{alg:occlusion} summarizes the complete rendering procedure.

\begin{figure}[!t]
  \centering
  \input{images_tex/cross.tex}
\end{figure}

\subsection{Video Diffusion with Dual-branch Conditioning}
\label{subsec:diffusion}

\subsubsection{Model Architecture}

For novel view synthesis, we extend the conditioning of the velocity field in Sec.~\ref{subsec:preliminary} to $\bm{c}=(\bm{y},\bm{I}^{r},\bm{M}^{r},\bm{I}^{s})$, comprising the text embedding $\bm{y}$, point cloud renders $\bm{I}^{r}$, validity masks $\bm{M}^{r}$, and source reference video $\bm{I}^{s}$.
While our preliminary version~\cite{mark2025trajectorycrafter} utilized the 5B-parameter CogVideoX~\cite{yang2024cogvideox}, we scale up our approach in this work by leveraging VACE~\cite{vace}, a 14B video generation and editing model.
To handle spatially aligned but incomplete geometry $\bm{I}^{r}$ and spatially misaligned source appearance $\bm{I}^{s}$, we adopt a dual-branch conditioning mechanism with a geometric-conditioning branch and a reference-injection branch, both connected to a shared video diffusion backbone.

The geometric-conditioning branch converts the occlusion-aware renderings into spatial guidance for target-view generation. It uses ControlNet-style~\cite{zhang2023adding} DiT blocks to process the rendered images $\bm{I}^{r}$ and validity masks $\bm{M}^{r}$, injecting the resulting conditioning features into the video diffusion backbone. We initialize these geometric-conditioning DiT (Geo-DiT) blocks from the context blocks of VACE~\cite{vace} and train them to guide generation using the available target-view structure.

The reference-injection branch complements geometric guidance by retrieving appearance information from the source video $\bm{I}^{s}$. Although the point cloud renders provide target-view spatial structure, they may lose fine texture details due to geometric inaccuracies and rendering artifacts. The source video preserves these details, but its spatial misalignment with the target views calls for reference conditioning that does not require pixelwise alignment. We therefore integrate reference-conditioned DiT (Ref-DiT) blocks into the video diffusion backbone, as illustrated in Fig.~\ref{fig:cross}. Within these blocks, cross-attention uses target-view features as queries ($Q$) and source-video features as keys ($K$) and values ($V$), allowing target-view features to retrieve relevant appearance information across source locations. This provides appearance cues for refining degraded textures and guiding completion in masked regions.

\input{images_tex/recon_view}

\subsubsection{Dataset Curation}

To train the model, we ideally require synchronized multi-view videos of diverse 4D scenes. However, existing multi-view datasets~\cite{corona2021meva, grauman2022ego4d, zheng2023pointodyssey, sener2022assembly101, greff2022kubric} tend to be small in scale, non-photorealistic, or lacking in diversity. Training solely on such datasets would limit performance and generalization in real-world scenarios.

Fortunately, our approach explicitly decouples view transformation from content generation, allowing us to curate training data specifically for the video diffusion model with dual-branch conditioning. We propose distinct strategies for processing web-scale monocular videos and static multi-view datasets.

For monocular datasets, we propose a self-supervised strategy by repurposing the triple reprojection mechanism.
Given a source video $\bm{I}^{s}$, we lift it into a dynamic point cloud $\bm{P}^{s}$ using Eq.~\ref{eq:pcd}. To simulate the geometric artifacts inherent in novel view synthesis, we apply a random relative view transformation $\Delta\bm{T}$ to render an intermediate view.
Subsequently, we back-project this intermediate view and re-render it back to the original pose using the inverse transformation $\Delta\bm{T}^{-1}$, yielding the degraded view $\bm{I}''$ and its corresponding validity mask $\bm{M}''$.
As shown in Fig.~\ref{fig:double_reprojection} (row 3), $\bm{I}''$ is spatially realigned with the ground truth $\bm{I}^{s}$ but contains holes and invalid regions characteristic of novel view rendering.
Consequently, we construct training pairs using the degraded reprojection and its mask $(\bm{I}'', \bm{M}'')$ as the geometric condition and the pristine source video $\bm{I}^{s}$ as supervision.
We adopt the OpenVid-1M~\cite{nan2024openvid} dataset and employ Video Depth Anything~\cite{vda} to generate 100,000 such pairs.

While web-scale monocular videos offer diverse dynamic content, they predominantly feature limited camera motion with small baselines.
To complement this and incorporate real-world large-baseline view changes, we leverage datasets of static scenes captured with extensive camera movements~\cite{ling2024dl3dv,zhou2018real10k}.
We design a pipeline to construct training triplets comprising a source video $\bm{I}^s$, a target ground-truth video $\bm{I}^t$, and a geometrically aligned point cloud render $\bm{I}^r$ and its mask $\bm{M}^r$.
Specifically, for each video sequence, we employ VGGT~\cite{wang2025vggt} to simultaneously reconstruct the global point cloud and estimate camera poses for all frames.
We then sample two clips with visual overlap from the sequence, designating one as the source input $\bm{I}^s$ and the other as the target ground truth $\bm{I}^t$.
Finally, we render the point cloud derived from $\bm{I}^s$ onto the camera trajectory of $\bm{I}^t$ to generate the geometric condition $\bm{I}^r$ and its mask $\bm{M}^r$.
Using the DL3DV~\cite{ling2024dl3dv} and RealEstate10K~\cite{zhou2018real10k} datasets, we generate 100,000 such static multi-view training samples.

\setlength{\dblfloatsep}{14pt plus 2pt minus 2pt}
\setlength{\dbltextfloatsep}{18pt plus 2pt minus 2pt}
\subsubsection{Training Scheme}
We adapt the conditional velocity field through a two-stage training strategy that first establishes geometric control and then learns appearance retrieval.
In the first stage, we train the Geo-DiT blocks in the geometric-conditioning branch on self-supervised pairs from dynamic monocular videos. The degraded reprojection and its validity mask provide the geometric condition, and the original video provides the reconstruction target. We keep the DiT backbone and the reference-injection branch frozen.
In the second stage, we freeze the trained Geo-DiT blocks and the backbone and train the Ref-DiT blocks in the reference-injection branch on static multi-view triplets. The source clip provides the appearance reference, while its point cloud is rendered along the target trajectory to supply geometric guidance. Supervision from the observed target clip teaches the branch to retrieve source appearance across viewpoints.

\subsection{Scene Generation}

Generating 3D or 4D scenes from monocular observations is challenging due to limited multi-view constraints. We address this by synthesizing synchronized videos from fixed novel viewpoints, providing additional supervision for scene reconstruction. Below, we focus on 4D reconstruction from monocular video, with single-image 3D reconstruction treated as a static special case.

\subsubsection{Layer-Wise Point Cloud Construction}
To provide consistent geometric conditions for multi-view video generation, we construct layered point clouds with a completed background and detailed foreground geometry. Novel viewpoints can expose background regions hidden by moving foreground objects, so we complete the background before generating new views. We initialize foreground segmentation on the first source frame $\bm{I}^{s}_{1}$ using image matting~\cite{CarveKit}, then use sampled foreground points to prompt SAM 2~\cite{sam2} for tracking across the video. Taking the union of the tracked masks yields the foreground mask sequence $\bm{M}^{\text{fg}}$. We remove the foreground and apply video inpainting~\cite{miao2025rose} to obtain a completed background video $\bm{I}^{\text{bg}}$. From this video, STream3R~\cite{lan2025stream3r} estimates background depth $\bm{D}^{\text{bg}}$ and camera parameters $[\bm{T}^{s},\bm{K}]$.

\begingroup
The foreground geometry must preserve fine details while remaining compatible with the background geometry. We use STream3R on the source video $\bm{I}^{s}$ to obtain a coarse depth estimate $\bm{D}^{\text{coarse}}$ as the geometric reference, and Video Depth Anything~\cite{vda} to obtain a detailed depth estimate $\bm{D}^{\text{fine}}$. We align the detailed estimate to the reference in inverse-depth space by fitting a scale and shift within the foreground mask for each frame. Temporal smoothing of these parameters reduces frame-to-frame variation, following a momentum-based strategy~\cite{huang2025vipe}:
\begin{equation}
  \begin{aligned}
    &\alpha_i, \beta_i = \operatorname*{argmin}_{\alpha, \beta}\,
    \Bigl\lVert \bm{M}_i^\text{fg} \cdot \bigl(\alpha / \bm{D}_i^\text{fine} + \beta - 1/\bm{D}_i^\text{coarse}\bigr) \Bigr\rVert_2^2, \\
    &\hat{\alpha}_i = m \cdot \hat{\alpha}_{i-1} + (1-m) \cdot \alpha_i, \\
    &\hat{\beta}_i = m \cdot \hat{\beta}_{i-1} + (1-m) \cdot \beta_i,
  \end{aligned}
\end{equation}
where $m$ is the momentum factor. The aligned foreground depth is $\bm{D}_i^\text{fg} = \bm{M}_i^\text{fg} \cdot (\hat{\alpha}_i / \bm{D}_i^\text{fine} + \hat{\beta}_i)^{-1}$.

We then combine the aligned foreground depth with the background depth using the foreground mask:
\begin{equation}
\bm{D}^\text{s} = \bm{M}^\text{fg} \cdot \bm{D}^\text{fg} + (1 - \bm{M}^\text{fg}) \cdot \bm{D}^\text{bg},
\label{eq:depth_comp}
\end{equation}
where the products, divisions, and reciprocals are applied elementwise. The foreground and completed background provide a sequence of dynamic point clouds $\bm{P}^{s} = \{\bm{P}_{i}^{s}\}_{i=1}^{n}$, supplying geometric conditions for the static-view and synchronized-video generation described next.
\endgroup

\input{tables/table_3dnvs}
\input{images_tex/3dnvs_comp}
\input{tables/table_4d_zeroshot}
\input{images_tex/4dnvs_comp_noref}
\input{images_tex/4dnvs_comp}
\input{tables/table_4d_scene_generation}
\input{images_tex/4dgen_comp}
\input{tables/table_worldscore}

\subsubsection{Multi-view Video Generation and 4D Reconstruction}

\begingroup
Our goal is to generate synchronized videos from multiple fixed cameras, providing observations of the dynamic scene from different viewpoints at each source time step. As illustrated by the diagonal sequence in Fig.~\ref{fig:view}(a), a video generated along a gradual camera trajectory changes viewpoint as time advances and therefore provides only one novel view per time step. To obtain multi-view observations throughout the sequence, we separate viewpoint changes from temporal evolution: we first establish the scene appearance across viewpoints at the initial time, then generate a dynamic video at each fixed viewpoint.

In the first stage, we freeze the scene at the initial source frame $\bm{I}_1^s$ and render its point cloud along the target camera trajectory. Conditioned on these renderings, our model synthesizes a sequence of novel views depicting the same scene state. Fig.~\ref{fig:view}(b) illustrates this fixed-time generation stage, in which only the viewpoint varies. The generated static views establish the initial appearance at the target viewpoints and serve as first-frame anchors for the dynamic videos generated in the next stage.

In the second stage, we iteratively generate dynamic videos at fixed target viewpoints, following the process illustrated in Fig.~\ref{fig:view}(c). At each iteration, we hold the selected camera pose fixed and render the dynamic point cloud over the source time steps to obtain geometric conditions. We replace the first rendered frame with the corresponding static view from the first stage to anchor the initial appearance. Our model then synthesizes the dynamic video conditioned on this rendering sequence and the source video $\bm{I}^s$. The generated video is subsequently used to update background visibility in the source point clouds, which are rendered to provide geometric conditions for the next viewpoint.

Finally, the generated videos provide additional views at the source time steps. We jointly use these videos and the source video to supervise the optimization of an explicit dynamic Gaussian scene representation~\cite{lei2024mosca}, achieving the transition from monocular observations to a 4D scene.
\endgroup

%% file: images_tex/double_reprojection.tex
\begin{figure}[!t]
  \centering
  \includegraphics[width=\columnwidth]{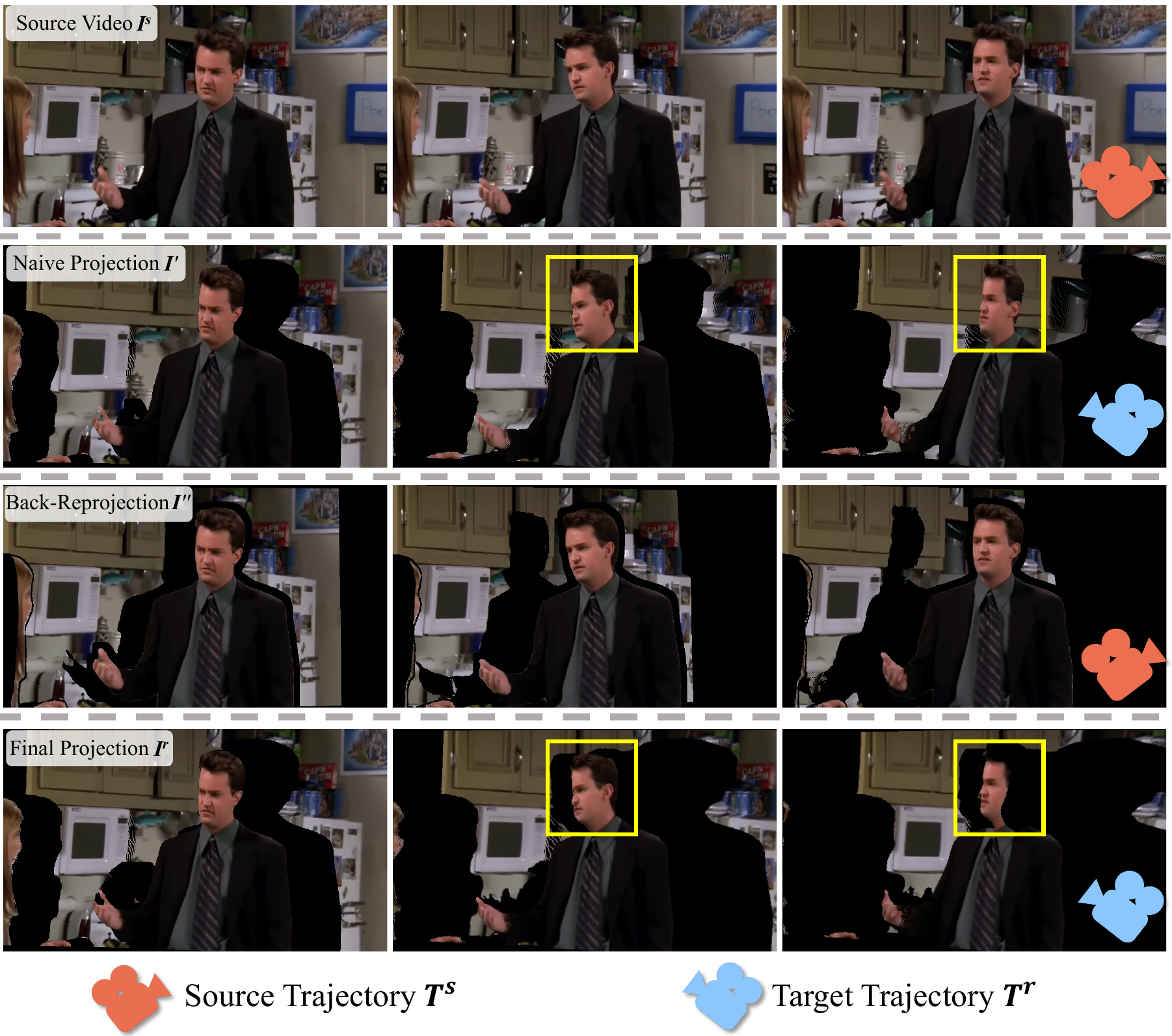}
  \caption{\textbf{Triple-reprojection-based occlusion disambiguation.}
  Starting from the source video $\bm{I}^{s}$ (first row), the first projection produces target-view renders $\bm{I}'$ with foreground--background tearing (second row). The second projection maps the intermediate geometry back to the source cameras, producing images $\bm{I}''$ (third row) and validity masks $\bm{M}''$. These masks are accumulated and combined with surface-orientation filtering to select source points for the third projection, yielding the final target-view renders $\bm{I}^{r}$ (fourth row). Yellow boxes highlight the reduction of erroneous projections; the remaining holes are left for subsequent video generation.
  }
\label{fig:double_reprojection}
\end{figure}

%% file: images_tex/normal.tex
\centering
\includegraphics[width=\linewidth]{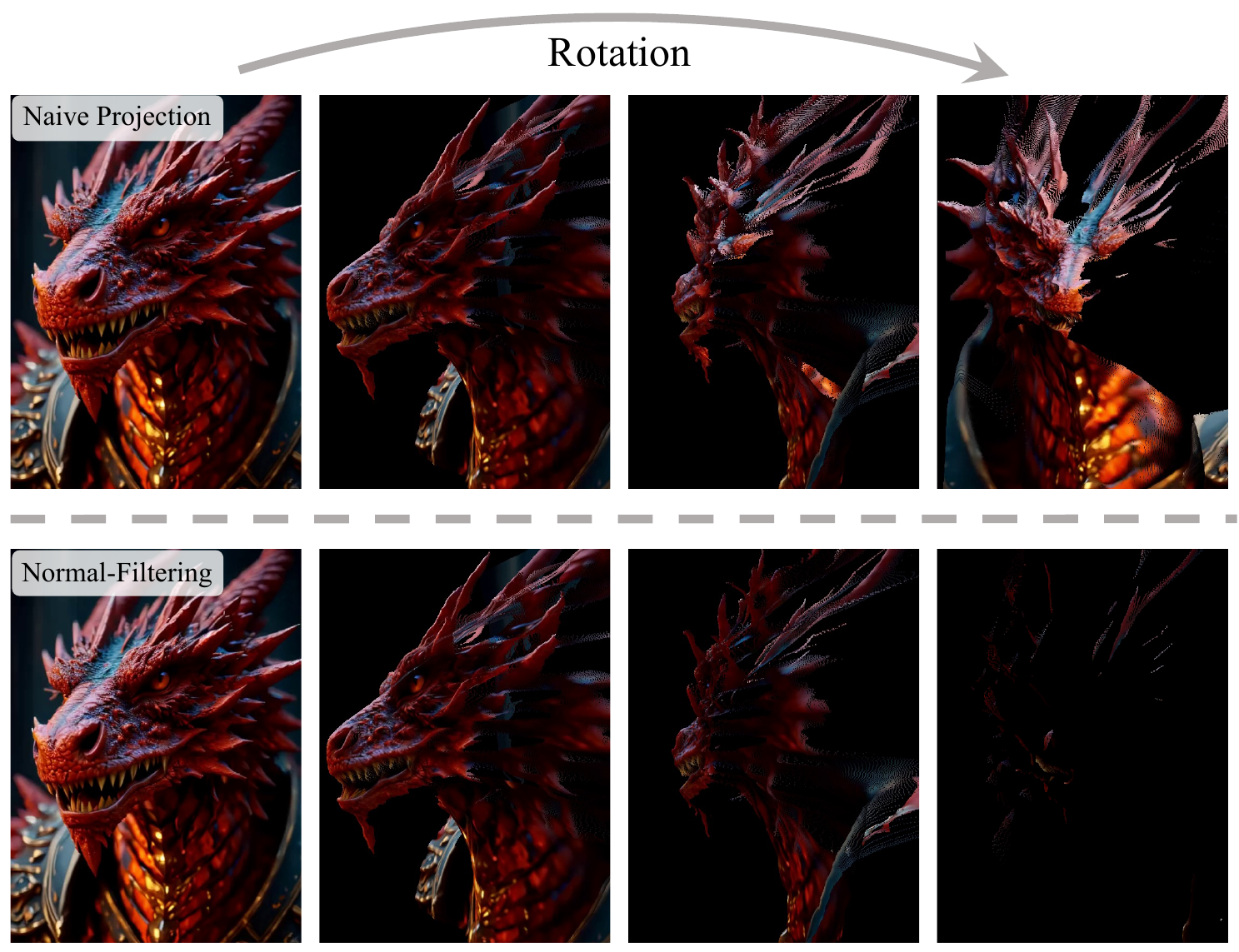}
\caption{\textbf{Normal-based visibility correction.}
Naive projection includes points on surfaces facing away from the target camera (back-faces), introducing geometric ambiguities. We filter out points whose surface normals deviate too much from the viewing direction.
}
\label{fig:normal}

%% file: images_tex/cross.tex
\centering
\includegraphics[width=0.85\linewidth]{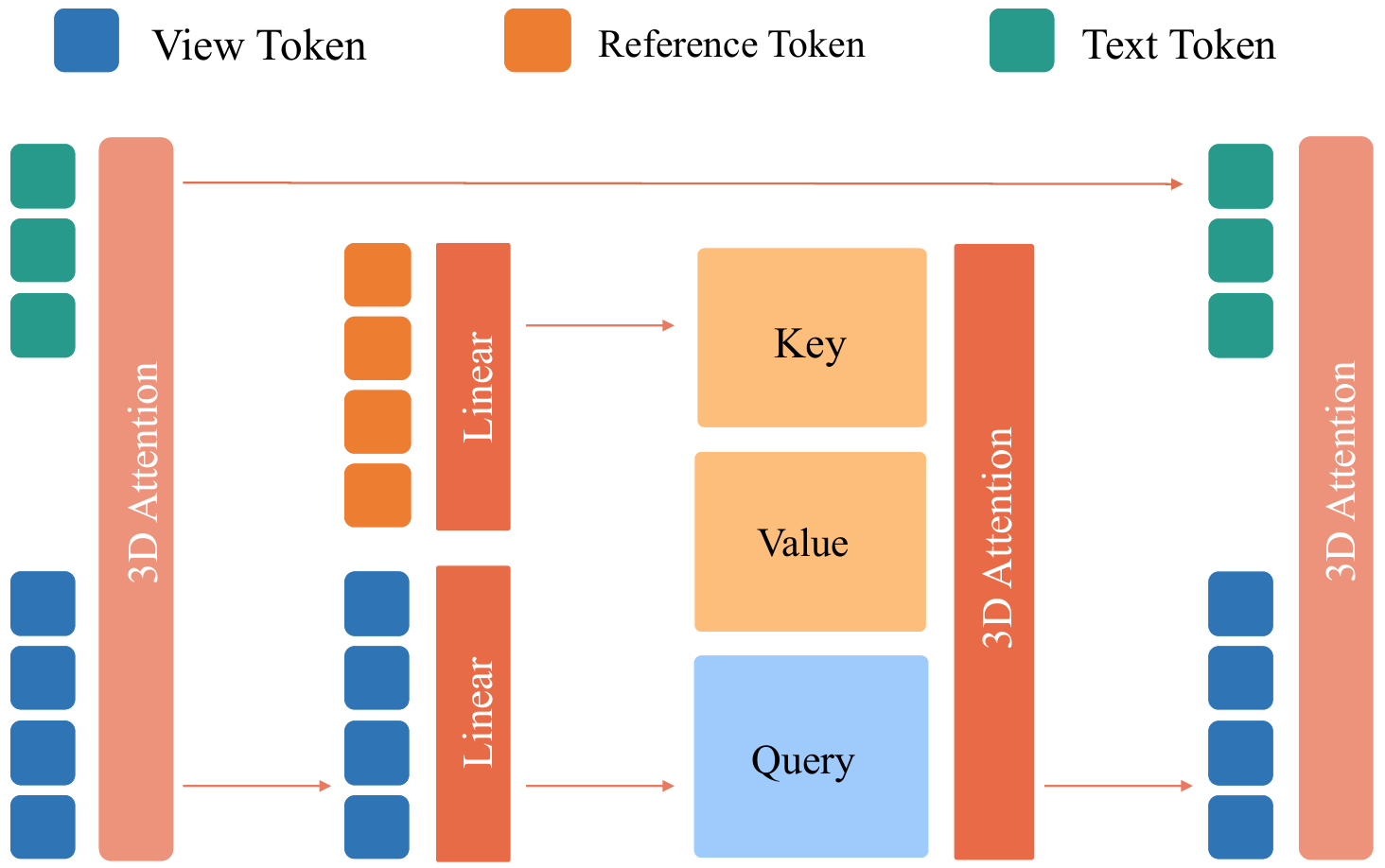}
\caption{\textbf{Ref-DiT block.} The text and view tokens are first processed by 3D attention, followed by cross-attention that injects detailed but spatially misaligned reference information into the view tokens, yielding refined features for subsequent layers.}
\label{fig:cross}

%% file: images_tex/recon_view.tex
\begin{figure}[!t]
  \centering
  \includegraphics[width=\linewidth]{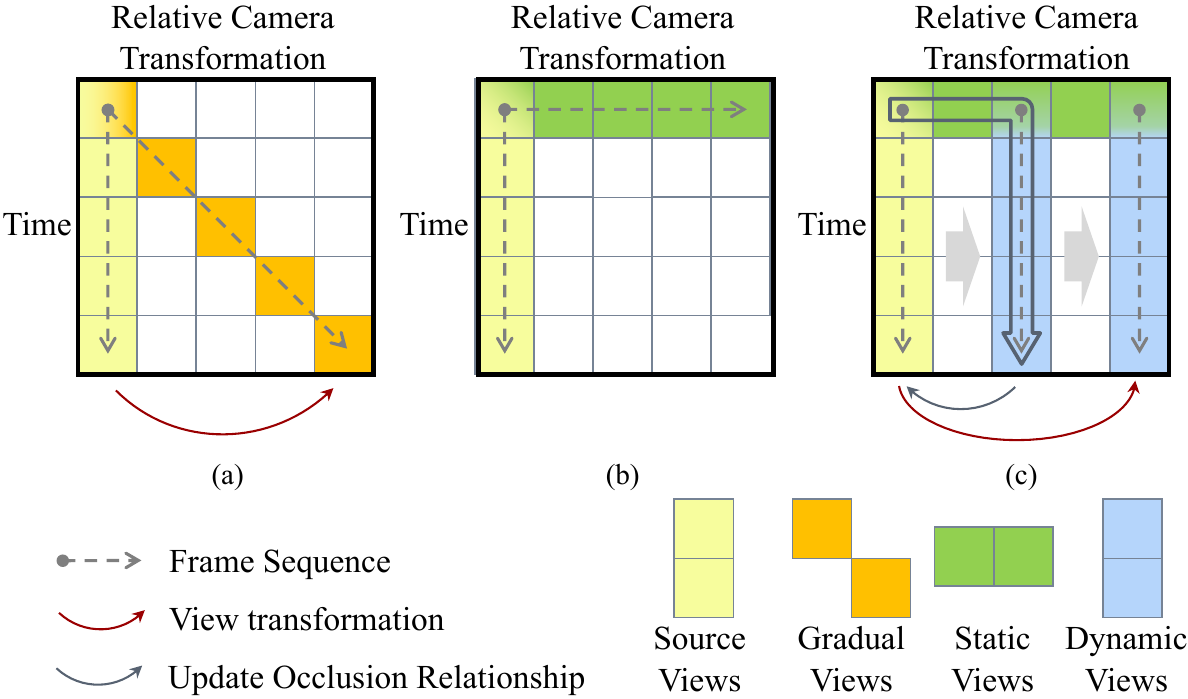}
  \caption{\textbf{Multi-view video generation for 4D reconstruction.}
  (a) Gradual camera trajectories change viewpoint and scene time together, tracing a diagonal through the grid. (b) We fix scene time at the initial source frame and vary the viewpoint to generate static views along the first row. (c) We then generate dynamic videos one viewpoint at a time, keeping the camera fixed while scene time varies along each column. The corresponding static view provides the first-frame appearance anchor. Foreground tracking follows the static views and the current dynamic video, and back-reprojection updates background visibility before generation proceeds to the next viewpoint. The synchronized fixed-camera videos provide multi-view supervision for 4D reconstruction.
  }
  \label{fig:view}
\end{figure}

%% file: tables/table_3dnvs.tex
\begin{table*}[t]
    \centering
    \tablebodyfont
    \setlength{\tabcolsep}{3pt}
    \caption{\textbf{Static scene novel view synthesis on CO3D~\cite{reizenstein21co3d}, RealEstate10K~\cite{zhou2018real10k}, and DL3DV~\cite{ling2024dl3dv}.} We evaluate each synthesized view against the corresponding ground-truth image. The \textbf{best} and \underline{second-best} scores are marked in each column.}
    \label{tab:setnvs_small}
    \vspace{-0.7em}
\begin{tabular*}{\textwidth}{@{\extracolsep{\fill}}lccccccccc@{}}
    \toprule
    \multirow{2}{*}{Method} & \multicolumn{3}{c}{CO3D} & \multicolumn{3}{c}{RealEstate10K} & \multicolumn{3}{c}{DL3DV} \\
    \cmidrule(lr){2-4} \cmidrule(lr){5-7} \cmidrule(lr){8-10}
     & PSNR $\uparrow$ & SSIM $\uparrow$ & LPIPS $\downarrow$ & PSNR $\uparrow$ & SSIM $\uparrow$ & LPIPS $\downarrow$ & PSNR $\uparrow$ & SSIM $\uparrow$ & LPIPS $\downarrow$ \\
    \midrule
    See3D~\cite{Ma2024See3D} & 16.2194 & 0.4590 & 0.4471 & 16.4622 & 0.5796 & 0.2932 & 13.4184 & 0.3828 & 0.4736 \\
    GEN3C~\cite{ren2025gen3c} & \underline{19.1098} & 0.5712 & 0.3773 & 21.4607 & 0.7461 & 0.1650 & 14.7240 & 0.4262 & 0.4843 \\
    Uni3C~\cite{uni3c} & {19.0275} & \underline{0.5752} & \underline{{0.3152}} & \underline{{21.5379}} & \underline{{0.7530}} & \textbf{{0.1508}} & \underline{{15.4094}} & \underline{{0.4303}} & \underline{0.3929} \\
    SEVA~\cite{seva} & {17.5807} & {0.4997} & {0.3845} & {17.0130} & {0.6081} & {0.2989} & {14.2962} & {0.4029} & {0.4443} \\
    \midrule
    \tablehighlight Ours & \textbf{19.9958} & \textbf{0.5787} & \textbf{0.3082} & \textbf{21.7261} & \textbf{0.7833} & \underline{0.1582} & \textbf{15.8190} & \textbf{0.4462} & \textbf{0.3915} \\
    \bottomrule
\end{tabular*}
\end{table*}

%% file: images_tex/3dnvs_comp.tex
\begin{figure*}[!t]
  \centering
  \includegraphics[width=1.\textwidth]{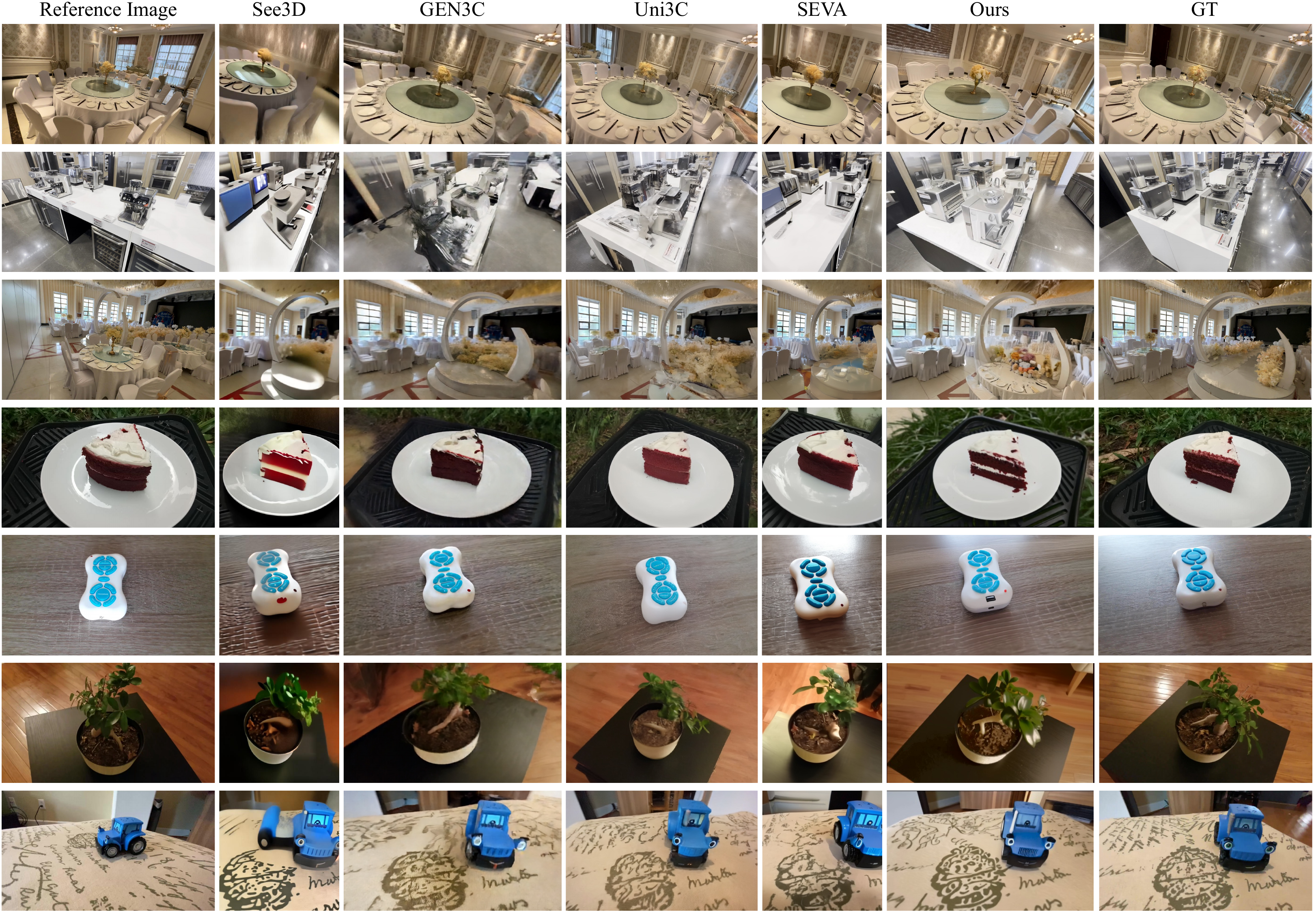}
  \caption{\textbf{Novel views synthesized from a single image of a static scene.} Each row shows the input image on the left, the views synthesized at the same target pose in the middle, and the ground-truth target view on the right.}
  \label{fig:3dnvs_comp}
\end{figure*}

%% file: tables/table_4d_zeroshot.tex
\begin{table*}[t]
    \centering
    \tablebodyfont
    \setlength{\tabcolsep}{3.8pt}
    \caption{\textbf{Dynamic scene novel view synthesis on iPhone and in-the-wild videos.} We measure target-view fidelity on five synchronized iPhone sequences, and evaluate camera accuracy and video quality on 70 in-the-wild videos. Translation errors are in centimeters and rotation errors in degrees. The \textbf{best} and \underline{second-best} scores are marked in each column.}
    \label{tab:4d_zeroshot}
    \vspace{-0.7em}
    \begin{tabular*}{\textwidth}{@{\extracolsep{\fill}}lcccccccccc@{}}
        \toprule
        & \multicolumn{3}{c}{Reference-based (iPhone)} & \multicolumn{7}{c}{Reference-free (In-the-Wild)} \\
        \cmidrule(lr){2-4} \cmidrule(lr){5-11}
        Method & \multirow{2}{*}[-2.2pt]{PSNR$\uparrow$} & \multirow{2}{*}[-2.2pt]{SSIM$\uparrow$} & \multirow{2}{*}[-2.2pt]{LPIPS$\downarrow$} & \multicolumn{2}{c}{APE} & \multicolumn{2}{c}{RPE $d{=}1$} & \multicolumn{3}{c}{Video quality} \\
        \cmidrule(lr){5-6} \cmidrule(lr){7-8} \cmidrule(lr){9-11}
        & & & & Trans.$\downarrow$ & Rot.$\downarrow$ & Trans.$\downarrow$ & Rot.$\downarrow$ & CLIP$\uparrow$ & FVD$\downarrow$ & FID$\downarrow$ \\
        \midrule
        TrajectoryCrafter~\cite{mark2025trajectorycrafter}
        & 13.60 & \underline{0.348} & 0.517 & \underline{1.56} & \underline{1.88} & \textbf{0.62} & 0.38 & \underline{0.946} & 614.0 & 47.9 \\
        ReCamMaster~\cite{bai2025recammaster}
        & 10.38 & 0.275 & 0.723 & 4.50 & 2.22 & 1.51 & \underline{0.27} & 0.893 & 774.7 & 69.1 \\
        EX-4D~\cite{hu2025ex4d}
        & 12.34 & 0.254 & 0.532 & 3.24 & 5.83 & 1.18 & 1.30 & 0.912 & 1048.8 & 75.9 \\
        WorldForge~\cite{song2025worldforge}
        & 11.68 & 0.315 & 0.563 & 3.42 & 3.65 & 1.00 & 0.88 & 0.933 & \underline{568.9} & 57.6 \\
        FreeOrbit4D~\cite{freeorbit4d}
        & 9.06 & 0.230 & 0.731 & 2.08 & 3.37 & 0.82 & 0.66 & 0.916 & 879.5 & 78.7 \\
        Reshoot-Anything~\cite{reshootanything}
        & \underline{13.76} & 0.343 & \textbf{0.468} & 4.20 & 5.89 & 0.85 & 1.62 & 0.944 & 569.7 & \textbf{44.2} \\
        \midrule
        \tablehighlight Ours
        & \textbf{14.44} & \textbf{0.364} & \underline{0.472} & \textbf{1.39} & \textbf{1.18} & \underline{0.70} & \textbf{0.14} & \textbf{0.952} & \textbf{546.3} & \underline{44.4} \\
        \bottomrule
    \end{tabular*}\end{table*}

%% file: images_tex/4dnvs_comp_noref.tex
\begin{figure*}[!t]
  \centering
  \includegraphics[width=\textwidth]{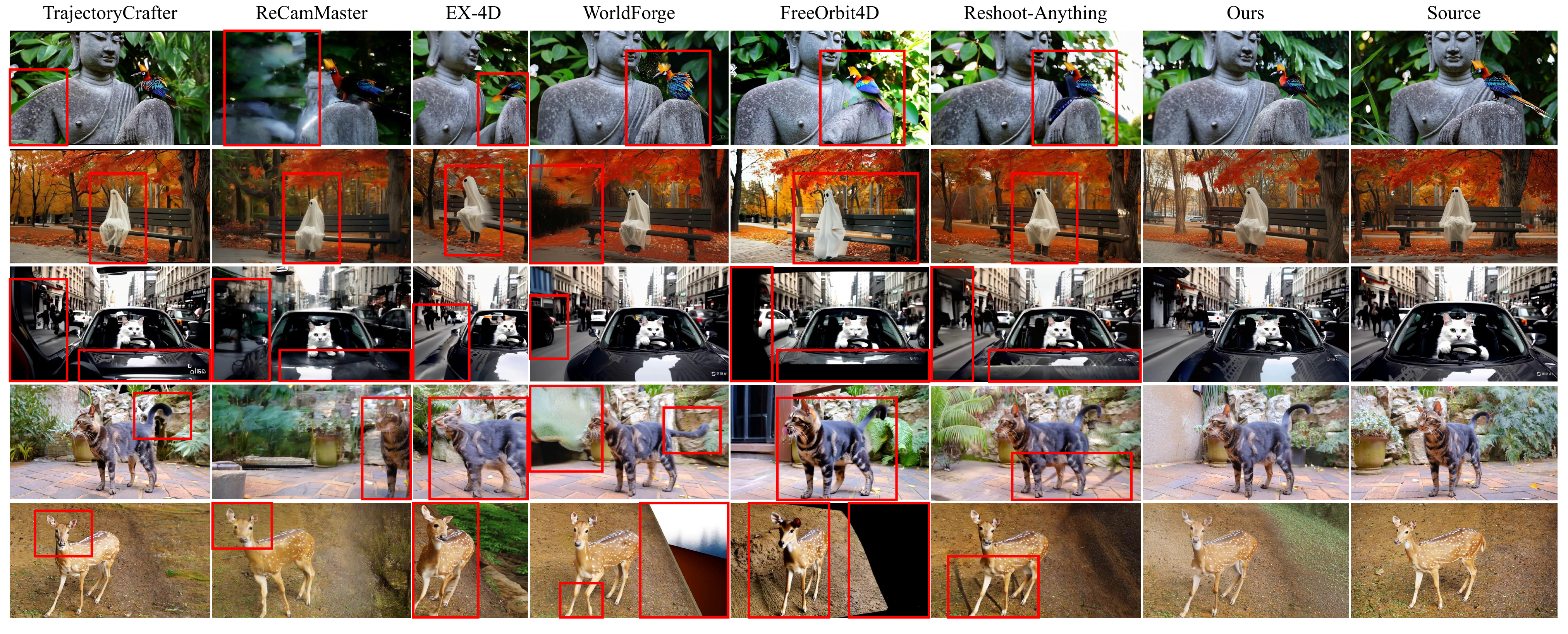}
  \caption{\textbf{Novel view synthesis on in-the-wild videos.} Each row compares frames synthesized for the same target camera pose. The rightmost column shows the source view for appearance reference. Red boxes mark regions for comparison.}
  \label{fig:4dnvs_comp_noref}
\end{figure*}

%% file: images_tex/4dnvs_comp.tex
\begin{figure*}[!t]
  \centering
  \includegraphics[width=\textwidth]{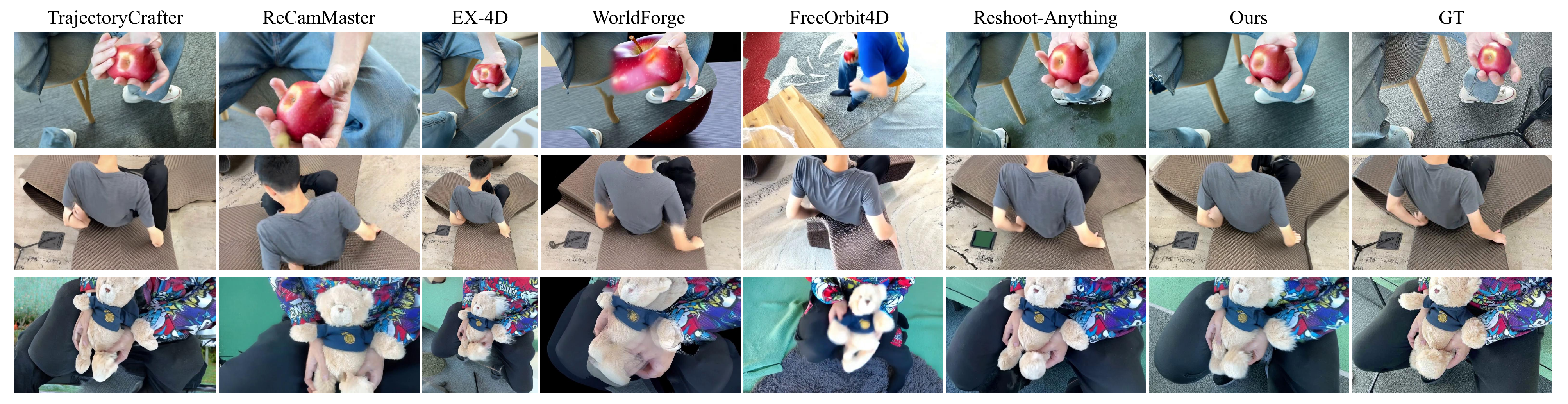}
  \caption{\textbf{Novel view synthesis on the iPhone dataset.} Each row compares views synthesized from a handheld video with the synchronized target frame from the fixed camera (rightmost column).}
  \label{fig:4dnvs_comp}
\end{figure*}

%% file: tables/table_4d_scene_generation.tex
\begin{table*}[t]
    \centering
    \tablebodyfont
    \setlength{\tabcolsep}{3pt}
    \caption{\textbf{4D scene generation from monocular video.} We render each of 13 reconstructed scenes from three fixed cameras. VBench evaluates the 39 rendered videos, while CLIP similarity, FVD, and FID use the source videos as references. The \textbf{best} and \underline{second-best} scores are marked in each column.}
    \label{tab:4d_scene_generation}
    \vspace{-0.7em}
    \begin{tabular*}{\textwidth}{@{\extracolsep{\fill}}lcccccccc@{}}
        \toprule
        \multirow{2}{*}{Method}
        & \multicolumn{5}{c}{VBench Metrics} & \multicolumn{3}{c}{Frame-based Metrics} \\
        \cmidrule(lr){2-6} \cmidrule(lr){7-9}
        & \shortstack{Subject\\Consistency$\uparrow$} & \shortstack{Background\\Consistency$\uparrow$} & \shortstack{Motion\\Smoothness$\uparrow$} & \shortstack{Aesthetic\\Quality$\uparrow$} & \shortstack{Imaging\\Quality$\uparrow$} & \shortstack{CLIP\\Similarity$\uparrow$} & \raisebox{0.55em}{FVD$\downarrow$} & \raisebox{0.55em}{FID$\downarrow$} \\
        \midrule
        Free4D~\cite{liu2025free4d}
        & 0.9466 & 0.9620 & 0.9836 & 0.4919 & 0.4539 & 0.8535 & 2118.6 & 136.2 \\
        DreamScene4D~\cite{dreamscene4d}
        & 0.9683 & \textbf{0.9853} & \textbf{0.9956} & 0.3511 & 0.5148 & 0.7286 & 2328.6 & 220.0 \\
        MoSca~\cite{lei2024mosca}
        & \underline{0.9685} & 0.9724 & 0.9927 & \underline{0.5265} & \underline{0.6663} & \underline{0.9392} & \underline{590.3} & \underline{76.9} \\
        \midrule
        \tablehighlight Ours
        & \textbf{0.9727} & \underline{0.9736} & \underline{0.9934} & \textbf{0.5580} & \textbf{0.6904} & \textbf{0.9523} & \textbf{553.0} & \textbf{50.0} \\
        \bottomrule
    \end{tabular*}\end{table*}

%% file: images_tex/4dgen_comp.tex
\begin{figure*}[!t]
  \centering
  \includegraphics[width=.9\textwidth]{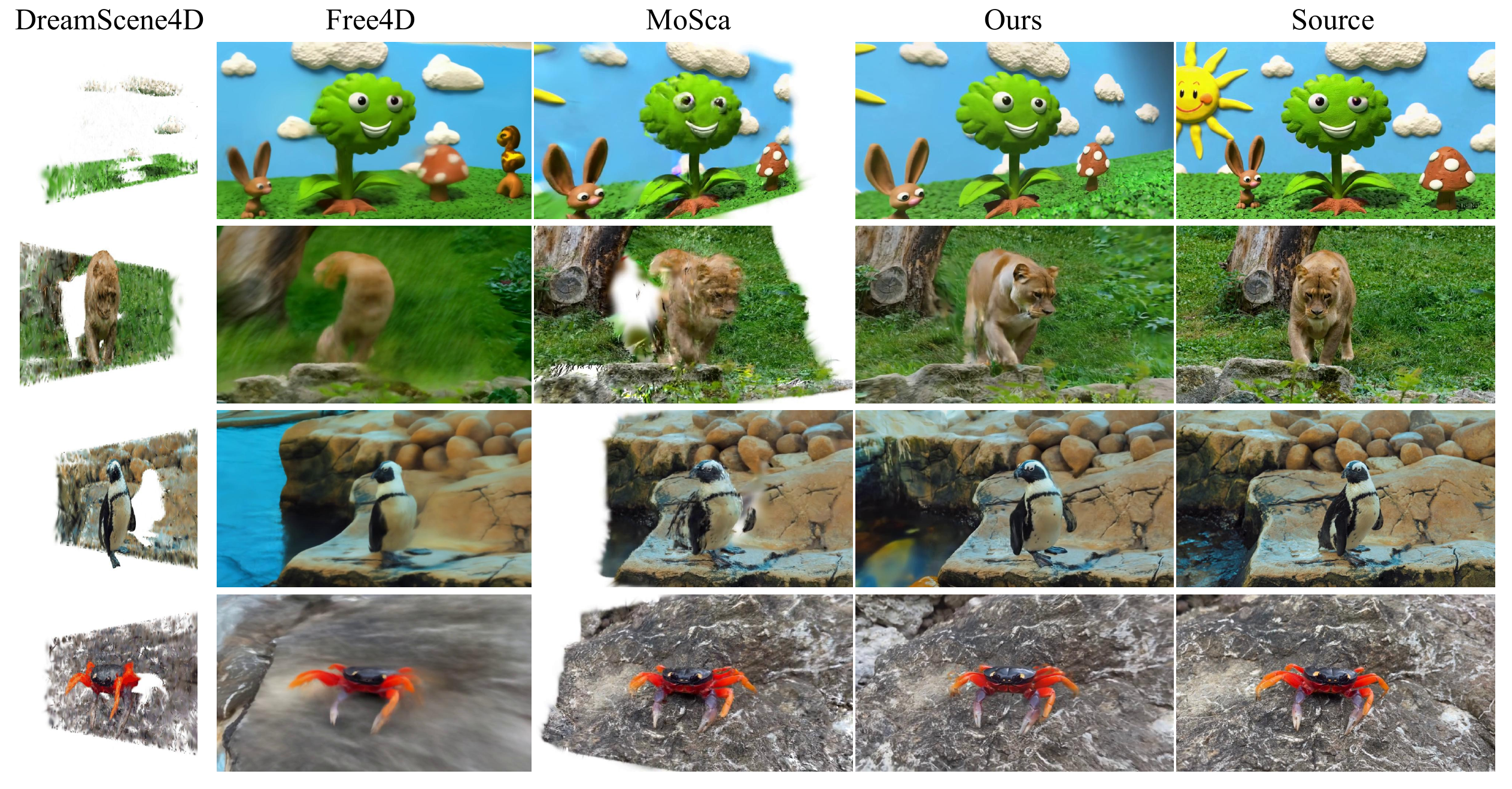}
  \caption{\textbf{Fixed-camera renderings of reconstructed 4D scenes.} Each row compares renderings from the same novel viewpoint at the same time step. The rightmost column shows the corresponding source frame. Full videos are provided in the supplementary material.}
  \label{fig:4dgen_comp}
\end{figure*}

%% file: tables/table_worldscore.tex
\begin{table*}[t]
    \centering
    \tablebodyfont
    \setlength{\tabcolsep}{2pt}
    \caption{\textbf{WorldScore leaderboard comparison.} We report our July 23, 2026 submission alongside the eight highest-ranked independent model families by WorldScore-Static, retaining the highest-ranked entry per family. S and D denote WorldScore-Static and WorldScore-Dynamic. The component scores are camera control (CC), object control (OC), content alignment (CA), 3D consistency (3DC), photometric consistency (PC), style consistency (SC), subjective quality (SQ), motion accuracy (MA), motion magnitude (MM), and motion smoothness (MS). Higher scores are better. The \textbf{best} and \underline{second-best} scores are marked in each column.}
    \label{tab:worldscore}
    \vspace{-0.7em}

    \begin{tabular*}{\textwidth}{@{\extracolsep{\fill}}l*{12}{c}@{}}
        \toprule
        & \multicolumn{2}{c}{WorldScore} & \multicolumn{3}{c}{Controllability}
        & \multicolumn{4}{c}{Quality and Consistency} & \multicolumn{3}{c}{Dynamics} \\
        \cmidrule(lr){2-3} \cmidrule(lr){4-6} \cmidrule(lr){7-10} \cmidrule(lr){11-13}
        Method & S & D & CC & OC & CA & 3DC & PC & SC & SQ & MA & MM & MS \\
        \midrule
        WorldScape-0.2 (MoE)~\cite{worldscapemoe2026} & \underline{85.13} & \textbf{76.23} & 94.32 & 87.65 & 78.13 & 86.51 & 90.02 & \underline{90.52} & \underline{68.75} & 61.21 & 23.06 & \underline{82.11} \\
        World Dreamer~\cite{worlddreamerentry2026} & 84.52 & 74.35 & 91.62 & 86.29 & 79.08 & 89.54 & 90.31 & 87.78 & 67.01 & 58.94 & 24.34 & 68.58 \\
        EvoPhys-World~\cite{evophysworld2026} & 83.45 & 73.78 & \underline{94.56} & 85.52 & \underline{83.37} & 86.78 & 85.56 & 83.24 & 65.12 & 62.47 & \underline{25.18} & 66.00 \\
        Inspatio-World~\cite{inspatioworld2026} & 83.15 & 74.10 & 93.51 & 84.67 & 75.26 & 86.35 & 89.40 & 85.79 & 67.07 & 52.07 & 24.15 & \textbf{82.71} \\
        EonWorld~\cite{eonworldentry2026} & 81.08 & 73.37 & 79.51 & 61.00 & 82.12 & \textbf{92.76} & 89.18 & \textbf{96.70} & 66.28 & \underline{65.35} & 24.75 & 76.81 \\
        FantasyWorld-1.0~\cite{dai2026fantasyworld} & 80.45 & 71.39 & 81.45 & \underline{87.90} & 66.94 & 84.62 & \underline{94.07} & 86.69 & 61.46 & 50.30 & 24.61 & 75.81 \\
        TeleWorld~\cite{chen2025teleworld} & 78.23 & 66.73 & 76.58 & 74.44 & 73.20 & 87.35 & 88.82 & 85.59 & 61.66 & 53.94 & \textbf{31.55} & 34.18 \\
        Voyager~\cite{huang2025voyager} & 77.62 & 54.53 & 85.95 & 66.92 & 68.92 & 81.56 & 85.99 & 84.89 & \textbf{71.09} & 0.00 & 0.00 & 0.00 \\
        \midrule
        \tablehighlight Ours & \textbf{85.53} & \underline{76.09} & \textbf{97.72} & \textbf{88.98} & \textbf{86.61} & \underline{91.63} & \textbf{94.11} & 76.55 & 63.12 & \textbf{77.66} & 24.12 & 60.42 \\
        \bottomrule
    \end{tabular*}
\end{table*}

%% file: sections/4_experiment.tex
\section{Experiments}
\label{sec:experiments}
\subsection{Implementation}
\label{subsec:implementation}

We build our video diffusion model with dual-branch conditioning on the pretrained VACE framework~\cite{vace}. All training videos are resized to $480\times832$ and sampled as 81-frame clips. Training proceeds in two stages. First, we optimize the Geo-DiT blocks on 100,000 self-supervised dynamic monocular pairs for 10,000 iterations with a learning rate of $1\times10^{-5}$, while freezing the DiT backbone. We then optimize the added Ref-DiT blocks on 100,000 static multi-view triplets for another 10,000 iterations with a learning rate of $5\times10^{-6}$, while keeping the remaining modules frozen. Both stages use a mini-batch size of 8 and are conducted on 32 GPUs. During data preprocessing, Video Depth Anything~\cite{vda} provides temporally consistent depth for monocular videos, while VGGT~\cite{wang2025vggt} reconstructs global point clouds and camera poses for static multi-view sequences. At inference time, a feed-forward geometry estimator recovers the source geometry, from which our occlusion-aware renderer produces geometric conditions for video generation.

For 4D scene generation, we adopt a two-stage view-generation strategy. In the first stage, we generate static novel-view sequences by progressively rotating the camera horizontally from the source pose to a user-defined target pose. These static views provide appearance anchors spanning the desired viewing range. In the second stage, we sample fixed camera poses at angular intervals of $5^{\circ}$ and iteratively generate synchronized dynamic videos over the pose intervals. The resulting dense multi-view videos are combined with the source video for 4D reconstruction~\cite{lei2024mosca}.

\subsection{Static Scene Novel View Synthesis}
\noindent\textbf{Datasets and evaluation metrics.}
We evaluate novel view synthesis for static scenes in a zero-shot setting on three real-world datasets spanning different scene scales: RealEstate10K~\cite{zhou2018real10k}, CO3D~\cite{reizenstein21co3d}, and DL3DV140 (DL3DV)~\cite{ling2024dl3dv}.
We follow the evaluation splits defined by SEVA~\cite{seva}.
Based on the disparity between input and target views, these splits cover two regimes: small-viewpoint NVS, which emphasizes smooth interpolation and temporal continuity, and large-viewpoint NVS, which tests the ability to synthesize substantial unseen content.
We report PSNR, SSIM, and LPIPS~\cite{zhang2018unreasonable} to evaluate image fidelity and perceptual quality.

\noindent\textbf{Comparison baselines.}
We compare against four generative NVS baselines: See3D~\cite{Ma2024See3D}, GEN3C~\cite{ren2025gen3c}, Uni3C~\cite{uni3c}, and SEVA~\cite{seva}.
See3D and SEVA rely on multi-view diffusion models, whereas GEN3C and Uni3C condition video diffusion models on explicit geometric priors.
See3D iteratively generates views along the target trajectory, using depth-based warping from previously synthesized views followed by multi-view diffusion inpainting.
GEN3C accepts an image, multiple views, or a video, reconstructs a point-cloud-based spatiotemporal cache, and conditions video generation on its novel-view renderings.
Uni3C augments the point-cloud cache with human motion and introduces a PCDController to jointly control the camera and human dynamics.
SEVA first generates keyframes along the target trajectory and then interpolates them sequentially.

\noindent\textbf{Qualitative comparison.}
Fig.~\ref{fig:3dnvs_comp} compares the methods under viewpoint changes. Existing approaches often distort object boundaries, produce implausible geometry, or degrade textures in regions newly revealed by the target camera. Our occlusion-aware renderer removes ambiguous projections and therefore provides more reliable geometric conditions, preserving scene structure and object contours under large-baseline transformations. The dual-branch conditioning architecture further separates geometric control from appearance retrieval, producing sharper textures and finer details.

\noindent\textbf{Quantitative comparison.}
As shown in Table~\ref{tab:setnvs_small}, \ourMethod{} achieves the best PSNR and SSIM on all three datasets, demonstrating strong zero-shot generalization across diverse real-world scenes. It also ranks first in LPIPS on CO3D and DL3DV, and second on RealEstate10K. Although Uni3C obtains lower LPIPS on RealEstate10K, \ourMethod{} provides a stronger overall balance among pixel fidelity, structural similarity, and perceptual quality. These results validate the effectiveness of our method.

\subsection{Dynamic Scene Novel View Synthesis}
\label{subsec:4d_zeroshot_nvs}

\subsubsection{Reference-Based Evaluation}
\noindent\textbf{Datasets and evaluation metrics.}
We evaluate novel view synthesis for dynamic scenes from monocular videos on the public iPhone dataset~\cite{iphone}. Following TrajectoryCrafter~\cite{mark2025trajectorycrafter}, we use five handheld sequences: Apple, Block, Paper-windmill, Spin, and Teddy. For each scene, the moving-camera video serves as the source, while the first synchronized fixed-camera clip provides target-view ground truth. COLMAP-derived camera poses and aligned depth maps from prior work~\cite{mark2025trajectorycrafter,wang2024shapeom} establish a shared geometric reference. We evaluate the synthesized target-view videos using PSNR, SSIM, and LPIPS~\cite{zhang2018unreasonable}.

\noindent\textbf{Comparison baselines.}
We compare against six diffusion-based methods for dynamic scene novel view synthesis: TrajectoryCrafter~\cite{mark2025trajectorycrafter}, ReCamMaster~\cite{bai2025recammaster}, EX-4D~\cite{hu2025ex4d}, WorldForge~\cite{song2025worldforge}, FreeOrbit4D~\cite{freeorbit4d}, and Reshoot-Anything~\cite{reshootanything}. TrajectoryCrafter conditions generation on point cloud renderings and masks, with a separate reference branch for appearance injection. ReCamMaster injects target-camera information implicitly into a source-video-conditioned diffusion model. EX-4D reconstructs a depth-watertight mesh and conditions an image-to-video model on its target-view renderings. WorldForge provides training-free trajectory control through inference-time latent guidance. FreeOrbit4D constructs a geometry-complete 4D proxy and combines rendered depth with the first source frame. Reshoot-Anything concatenates tokens from the source video and a 4D point cloud anchor video for self-attention-based generation. We evaluate all methods under the same protocol.

\noindent\textbf{Qualitative comparison.}
Fig.~\ref{fig:4dnvs_comp} shows target-view frames from three iPhone sequences under large viewpoint changes. Competing methods frequently distort foreground objects, destabilize scene structure, or introduce artifacts in newly exposed regions. ReCamMaster's implicit camera conditioning can become unstable in the presence of substantial source-video motion.
WorldForge lacks an explicit scene representation for enforcing geometry, while FreeOrbit4D can suffer from inconsistencies between rendered depth and its first-frame appearance reference. Reshoot-Anything directly mixes spatially misaligned source and target-view tokens, which can create ambiguous correspondences and background artifacts.

In contrast, our occlusion-aware renderer removes unreliable projections, including foreground--background tearing and back-facing surfaces, to produce cleaner geometric conditions and masks. Our dual-branch design then decouples structure from appearance, resulting in more stable camera control, coherent structures, and faithful textures.

\noindent\textbf{Quantitative comparison.}
Table~\ref{tab:4d_zeroshot} reports results averaged over the five scenes. \ourMethod{} achieves the best PSNR and SSIM, while ranking second in LPIPS. Overall, our occlusion-aware rendering and dual-branch conditioning improve geometric and photometric fidelity while maintaining competitive perceptual quality.

\subsubsection{Reference-Free Evaluation}
\noindent\textbf{Datasets and evaluation metrics.}
We collect an in-the-wild benchmark of 70 real and synthetic videos. The benchmark contains four categories of test trajectories with large camera translations or rotations. Following the reference-free evaluation protocol of WorldForge~\cite{song2025worldforge}, we compute FVD and FID between the source and generated target-view videos to assess overall video quality, together with frame-wise CLIP similarity to measure source-view consistency.

To evaluate camera-control accuracy, we use ViPE~\cite{huang2025vipe} to estimate the generated-video camera trajectory and compare it with the prescribed ground-truth trajectory. For APE, we first align the two trajectories using a $\mathrm{Sim}(3)$ transformation and then report translation-distance and rotation-angle errors. For RPE with $d{=}1$, we compare the relative poses of consecutive frames.

\noindent\textbf{Qualitative comparison.}
Fig.~\ref{fig:4dnvs_comp_noref} presents representative results on diverse real and synthetic videos. Although target-view ground truth is unavailable, comparison with the source frames reveals that existing methods often blur or omit foreground content, deform object shapes, or generate implausible background structures under large camera motions. \ourMethod{} better preserves salient subjects, scene layout, and texture coherence, demonstrating stronger generalization to unconstrained inputs.

\noindent\textbf{Quantitative comparison.}
As shown in Table~\ref{tab:4d_zeroshot}, \ourMethod{} achieves the lowest APE translation and rotation errors, the lowest RPE rotation error, and the best CLIP and FVD scores. It also ranks second in RPE translation error and FID. These results demonstrate accurate camera control and strong visual quality under large camera motions, even without target-view references.

\subsection{4D Scene Generation}
\label{subsec:4d_scene_generation}

\noindent\textbf{Datasets and evaluation metrics.}
Given a dynamic monocular video, our goal is to reconstruct a 4D scene representation that supports temporally consistent free-view rendering. Rather than optimizing solely from the observed input views, \ourMethod{} synthesizes dense, synchronized novel-view videos at a set of fixed camera poses. These videos provide appearance and geometric evidence for previously unseen regions, yielding stronger multi-view constraints for optimizing a dynamic 3DGS representation.

We select 13 dynamic videos from our in-the-wild benchmark. These casually captured videos contain diverse real-world motions and camera movements, providing a challenging testbed for complete 4D scene reconstruction from monocular observations.

For each source video, we reconstruct a 4D scene and render videos from three fixed camera poses, producing 39 test videos in total. We use VBench~\cite{vbench} to assess temporal consistency, motion smoothness, and perceptual quality. We additionally compute CLIP similarity, FVD, and FID between the rendered and source videos to measure reference consistency.

\noindent\textbf{Comparison baselines.}
We compare against two monocular-video 4D scene-generation methods based on dynamic 3D Gaussian Splatting: DreamScene4D~\cite{dreamscene4d} and Free4D~\cite{liu2025free4d}. DreamScene4D decomposes the video into dynamic objects and background components and reconstructs them through a decompose--recompose pipeline. Free4D applies a static novel-view synthesis model frame by frame to augment reconstruction with additional viewpoints. We also include MoSca~\cite{lei2024mosca}, our reconstruction backbone applied directly to the input video, to isolate the contribution of the dense novel views generated by \ourMethod{}.

\input{sections/renderer_figures}

\noindent\textbf{Qualitative comparison.}
Fig.~\ref{fig:4dgen_comp} presents fixed-view renderings of the reconstructed dynamic Gaussian scenes. DreamScene4D often leaves large unobserved regions incomplete, while Free4D and MoSca may blur dynamic subjects or destabilize scene structure away from the input trajectory. By augmenting reconstruction with dense synchronized views, \ourMethod{} preserves sharper foreground details and more coherent scene layouts while remaining visually consistent with the source video.

\noindent\textbf{Quantitative comparison.}
As shown in Table~\ref{tab:4d_scene_generation}, \ourMethod{} achieves the best subject consistency, aesthetic quality, imaging quality, CLIP similarity, FVD, and FID. It ranks second in background consistency and motion smoothness. The consistent improvements of \ourMethod{} over the MoSca backbone confirm that the synthesized dense views provide effective multi-view supervision for 4D reconstruction.

\begin{figure}[!t]
\centering
    \includegraphics[width=\columnwidth]{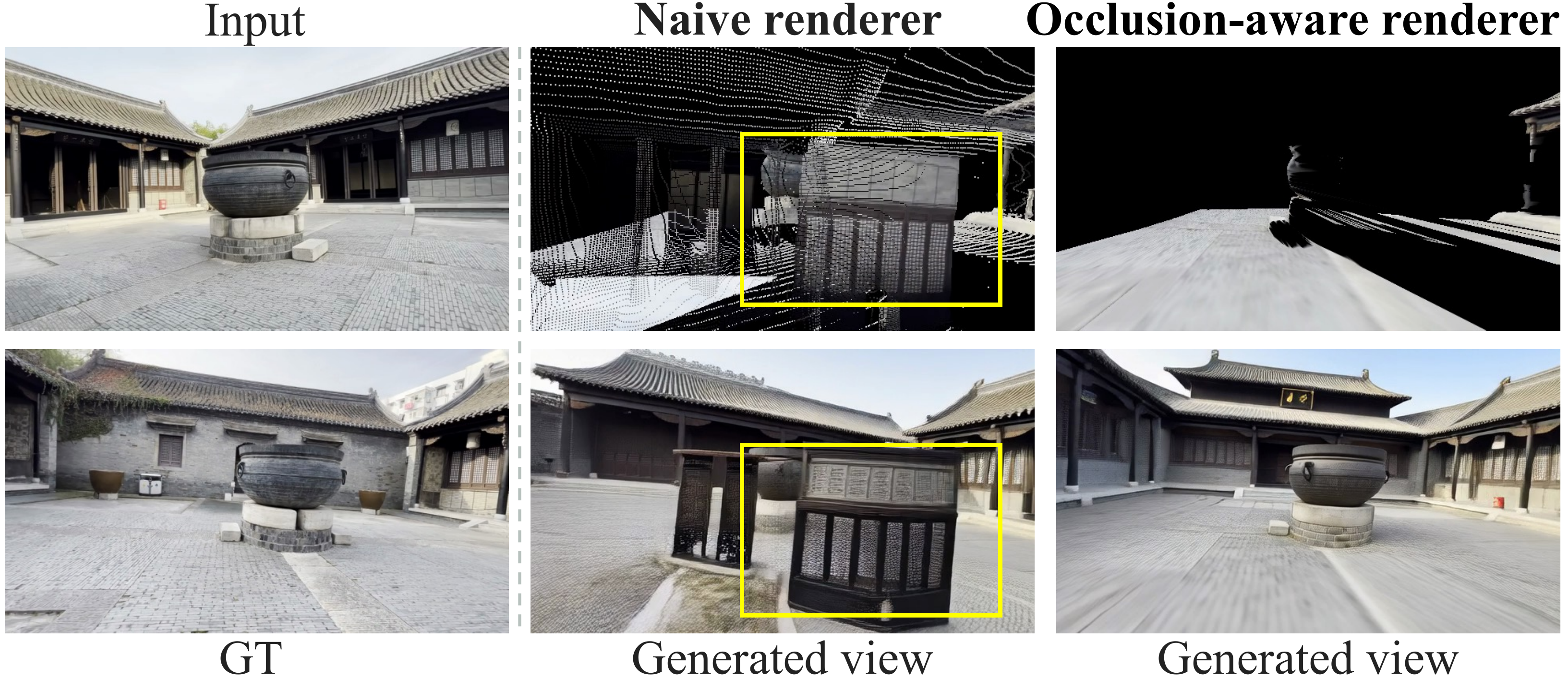}
    \caption{\textbf{Effect of the point cloud renderer on novel view synthesis.} The left column shows the input and target reference. The other columns pair each renderer's geometric condition (top) with the generated view (bottom). Yellow boxes mark projection artifacts that persist in the generated view.}
    \label{fig:renderer_qualitative}
\end{figure}

\subsection{Evaluation on WorldScore Benchmark}
We evaluate \ourMethod{} on the WorldScore benchmark{\hypersetup{linkcolor=black,citecolor=black}~\cite{duan2025worldscore}}\footnote{\url{https://huggingface.co/spaces/Howieeeee/WorldScore_Leaderboard}}. Each test case provides an initial frame, a text prompt, and a camera trajectory that specifies the desired scene layout. Table~\ref{tab:worldscore} reports WorldScore-Static (S) and WorldScore-Dynamic (D), together with camera control (CC), object control (OC), and content alignment (CA) for controllability; 3D consistency (3DC), photometric consistency (PC), style consistency (SC), and subjective quality (SQ) for quality and consistency; and motion accuracy (MA), motion magnitude (MM), and motion smoothness (MS) for dynamics. As shown in Table~\ref{tab:worldscore}, \ourMethod{} achieves the highest overall WorldScore-Static score and leads all three controllability metrics, while also performing strongly in 3D and photometric consistency. For dynamic-scene generation, we use a two-stage procedure. We first condition the model on the initial frame and text prompt to generate a dynamic reference video, and then combine this video with the prescribed camera trajectory to produce the final result. \ourMethod{} obtains the second-highest WorldScore-Dynamic score, demonstrating effective camera and content control for world generation.

\subsection{Ablation Study}
\label{subsec:ablation_study}

\noindent\textbf{Effectiveness of occlusion-aware point cloud renderer.}
We test whether our occlusion-aware point cloud renderer improves novel view synthesis over the direct projection used in prior work~\cite{yu2024viewcrafter,mark2025trajectorycrafter}. Both variants use the same source point cloud, target camera poses, and generation model. The baseline projects the source points directly into each target view. Our renderer uses triple reprojection to disambiguate occlusions and applies normal-based visibility filtering before the final projection to exclude points seen from the back or at excessively oblique angles (Algorithm~\ref{alg:occlusion}). To evaluate both variants over a controlled range of viewpoints, we reconstruct 30 scenes from the DL3DV validation split~\cite{ling2024dl3dv} using 3D Gaussian splatting (3DGS)~\cite{kerbl20233dgs}. For each scene, the 3DGS rendering at $0^\circ$ serves as the single input image. We evaluate the synthesized views at $30^\circ$ intervals up to $180^\circ$ against 3DGS renderings from the corresponding camera poses and average each metric over the 30 scenes.

As shown in Fig.~\ref{fig:renderer_angle}, the two renderers perform similarly at $30^\circ$, where the source and target views share much of the same visible surface area and direct projection already provides reliable guidance. As the angular baseline increases, more surfaces come into view and source points near occlusion boundaries can project onto the wrong surfaces. The benefit of filtering these projections is clearest from $120^\circ$ to $180^\circ$, where our renderer achieves significantly higher PSNR and SSIM and lower LPIPS at all three evaluated angles.

As shown in Fig.~\ref{fig:renderer_qualitative}, direct projection can incorrectly place background points and points near occlusion boundaries in foreground regions, introducing artifacts that carry over into the generated view. Our occlusion-aware renderer filters out these unreliable projections and better preserves foreground structure under large viewpoint changes.

\input{sections/ablation_floats}

\noindent\textbf{Ablation on model design.}
We ablate the two conditioning branches using the five-scene iPhone evaluation protocol from Sec.~\ref{subsec:4d_zeroshot_nvs} (Table~\ref{tab:ablation_model}). All three configurations receive the same target-view renders and validity masks. The \emph{w/o~Ref-DiT} variant retains Geo-DiT but removes the reference-injection branch. It can therefore recover source appearance only to the extent that the point cloud renders preserve it. The \emph{w/o~Geo-DiT} variant retains direct access to the source video through Ref-DiT. Following TrajectoryCrafter~\cite{mark2025trajectorycrafter}, it concatenates the encoded renders and masks with the noisy target latent at the input to the main DiT. This comparison isolates how the model incorporates the geometric conditions.

The full model performs best overall in Table~\ref{tab:ablation_model}. The two comparisons address complementary aspects of conditioning. With Geo-DiT in place, Ref-DiT gives target-view features access to source details even when projection has degraded their texture. With Ref-DiT in place, Geo-DiT supplies spatially aligned guidance at multiple backbone layers, beyond the guidance available from concatenation at the input.

\noindent\textbf{Ablation on 4D reconstruction.}
We compare two ways of generating novel views to supervise 4D reconstruction. In the \emph{Gradual} setting of Fig.~\ref{fig:view}(a), the camera moves along a single trajectory as the scene evolves, providing one generated viewpoint per time step. Our two-stage pipeline first synthesizes multiple views at the initial scene time (Fig.~\ref{fig:view}(b)), then uses each view to anchor a dynamic video at its fixed camera pose (Fig.~\ref{fig:view}(c)). This provides synchronized observations of the same motion from multiple viewpoints. We use the same evaluation setup as the 4D scene generation experiment in Sec.~\ref{subsec:4d_scene_generation}.

The two-stage pipeline achieves higher CLIP similarity and lower FVD and FID than the gradual baseline (Table~\ref{tab:ablation_reconstruction}). As shown in Fig.~\ref{fig:ablation_4dgs}, two-stage supervision also reduces visible artifacts in the reconstructed subject.

\begin{table}[!t]
\centering
    \tablebodyfont
    \caption{\textbf{Effect of novel-view supervision on 4D reconstruction.} We compare supervision from a video along a single camera trajectory (Gradual) with synchronized videos from multiple fixed cameras (two-stage) under the evaluation setup of Table~\ref{tab:4d_scene_generation}.}
    \label{tab:ablation_reconstruction}
    \begin{tabular*}{\columnwidth}{@{\extracolsep{\fill}}lccc@{}}
        \toprule
        Setting & \shortstack{CLIP\\Similarity$\uparrow$} & FVD$\downarrow$ & FID$\downarrow$ \\
        \midrule
        Gradual & 0.9171 & 740.5 & 93.3 \\
        Full two-stage pipeline & 0.9523 & 553.0 & 50.0 \\
        \bottomrule
    \end{tabular*}
\end{table}

%% file: sections/renderer_figures.tex
\begin{figure}[!t]
\makeatletter
\begingroup
\def\@captype{table}
\centering
    \tablebodyfont
    \caption{\textbf{Ablation of the conditioning branches on five iPhone scenes.} We remove Ref-DiT or replace Geo-DiT with channel-wise concatenation of the geometric conditions at the backbone input.}
    \label{tab:ablation_model}
    \setlength{\textwidth}{\columnwidth}
    \begin{tabular*}{\columnwidth}{@{\extracolsep{\fill}}lccc@{}}
        \toprule
        Setting & PSNR$\uparrow$ & SSIM$\uparrow$ & LPIPS$\downarrow$ \\
        \midrule
        w/o~Ref-DiT & 13.82 & 0.351 & 0.520 \\
        w/o~Geo-DiT & 13.37 & 0.350 & 0.528 \\
        \tablehighlight Full model & 14.44 & 0.364 & 0.472 \\
        \bottomrule
    \end{tabular*}
\par\endgroup
\makeatother
\medskip
\centering
    \includegraphics[width=\columnwidth]{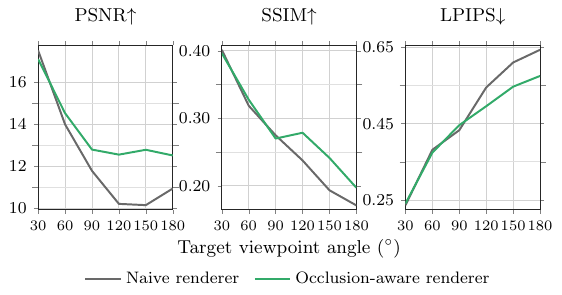}
    \caption{\textbf{Effect of the point cloud renderer across angular baselines.} Both renderers condition the same novel view synthesis model. At each angle, scores are averaged over 30 DL3DV validation scenes against target views rendered with 3D Gaussian splatting.}
    \label{fig:renderer_angle}
\end{figure}

%% file: sections/ablation_floats.tex
\begin{figure}[!t]
\centering
    \includegraphics[width=.9\columnwidth]{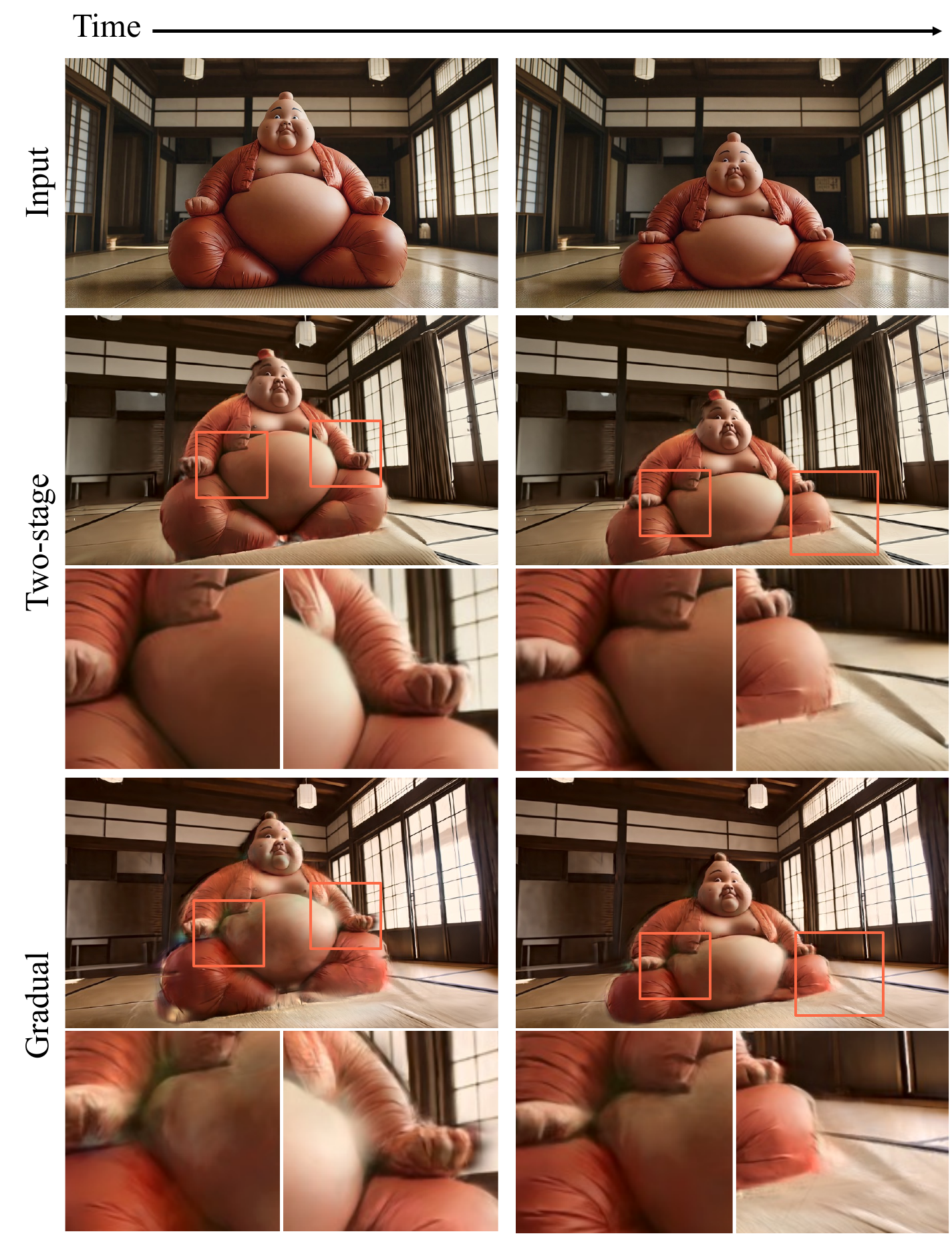}
    \caption{\textbf{Effect of novel-view supervision on reconstructed 4D scenes.} Columns show two time steps. Below the input frames, we show reconstructions supervised by synchronized fixed-camera videos (middle) and by a video along the gradual trajectory in Fig.~\ref{fig:view}(a) (bottom). Insets enlarge the boxed regions.}
    \label{fig:ablation_4dgs}
\end{figure}

%% file: sections/5_conclusion.tex
\section{Conclusion}

We presented \ourMethod{}, a unified framework for large-baseline novel view synthesis and scene generation from monocular images and videos. Our occlusion-aware point cloud renderer filters unreliable projections, and our dual-branch conditioning mechanism combines the resulting geometric guidance with appearance details from the source. Experiments on static and dynamic scenes show that this design enables the video diffusion model to follow a prescribed camera trajectory while preserving scene appearance. For 4D scene generation, our two-stage view-generation strategy separates viewpoint changes from scene motion to synthesize synchronized videos from multiple fixed cameras. Together with the source video, these novel views provide multi-view supervision for reconstructing an explicit dynamic Gaussian scene.

%% file: main.bib
@String(CVPR = {CVPR})

@String(ICCV = {ICCV})

@String(ECCV = {ECCV})

@String(NIPS = {NeurIPS})

@String(ICLR = {ICLR})

@String(SIGGRAPH = {SIGGRAPH})

@String(SIGGRAPHASIA = {SIGGRAPH Asia})

@String(PAMI = {IEEE TPAMI})

@String(TVCG = {IEEE TVCG})

@String(TOG = {ACM TOG})

@String(CGF = {Comput. Graph. Forum})

@String(AAAI = {AAAI})

@String(WACV = {WACV})

@String(CVPRW = {CVPR Workshops})

@inproceedings{lipman2023flow,
  title = {{Flow Matching for Generative Modeling}},
  author = {Lipman, Yaron and Chen, Ricky T. Q. and Ben-Hamu, Heli and Nickel, Maximilian and Le, Matthew},
  booktitle = ICLR,
  year = {2023}
}

@inproceedings{mildenhall2020nerf,
  title = {{NeRF: Representing Scenes as Neural Radiance Fields for View Synthesis}},
  author = {Mildenhall, Ben and Srinivasan, Pratul P. and Tancik, Matthew and Barron, Jonathan T. and Ramamoorthi, Ravi and Ng, Ren},
  booktitle = ECCV,
  year = {2020}
}

@inproceedings{DynNeRF,
  title = {{Dynamic View Synthesis From Dynamic Monocular Video}},
  author = {Gao, Chen and Saraf, Ayush and Kopf, Johannes and Huang, Jia-Bin},
  booktitle = ICCV,
  year = {2021}
}

@article{kerbl20233dgs,
  title = {{3D Gaussian Splatting for Real-Time Radiance Field Rendering}},
  author = {Kerbl, Bernhard and Kopanas, Georgios and Leimkuehler, Thomas and Drettakis, George},
  journal = TOG,
  year = {2023}
}

@inproceedings{4dgswu,
  title = {{4D Gaussian Splatting for Real-Time Dynamic Scene Rendering}},
  author = {Wu, Guanjun and Yi, Taoran and Fang, Jiemin and Xie, Lingxi and Zhang, Xiaopeng and Wei, Wei and Liu, Wenyu and Tian, Qi and Wang, Xinggang},
  booktitle = CVPR,
  year = {2024}
}

@inproceedings{wang2024shapeom,
  title = {{Shape of Motion: 4D Reconstruction from a Single Video}},
  author = {Wang, Qianqian and Ye, Vickie and Gao, Hang and Zeng, Weijia and Austin, Jake and Li, Zhengqi and Kanazawa, Angjoo},
  booktitle = ICCV,
  year = {2025}
}

@inproceedings{rombach2022high,
  title = {{High-Resolution Image Synthesis With Latent Diffusion Models}},
  author = {Rombach, Robin and Blattmann, Andreas and Lorenz, Dominik and Esser, Patrick and Ommer, Bj\"orn},
  booktitle = CVPR,
  year = {2022}
}

@article{chen2023videocrafter1,
  title = {{VideoCrafter1: Open Diffusion Models for High-Quality Video Generation}},
  author = {Chen, Haoxin and Xia, Menghan and He, Yingqing and Zhang, Yong and Cun, Xiaodong and Yang, Shaoshu and Xing, Jinbo and Liu, Yaofang and Chen, Qifeng and Wang, Xintao and Weng, Chao and Shan, Ying},
  journal = {arXiv:2310.19512},
  year = {2023}
}

@article{blattmann2023svd,
  title = {{Stable Video Diffusion: Scaling Latent Video Diffusion Models to Large Datasets}},
  author = {Blattmann, Andreas and Dockhorn, Tim and Kulal, Sumith and Mendelevitch, Daniel and Kilian, Maciej and Lorenz, Dominik and Levi, Yam and English, Zion and Voleti, Vikram and Letts, Adam and Jampani, Varun and Rombach, Robin},
  journal = {arXiv:2311.15127},
  year = {2023}
}

@inproceedings{blattmann2023align,
  title = {{Align Your Latents: High-Resolution Video Synthesis With Latent Diffusion Models}},
  author = {Blattmann, Andreas and Rombach, Robin and Ling, Huan and Dockhorn, Tim and Kim, Seung Wook and Fidler, Sanja and Kreis, Karsten},
  booktitle = CVPR,
  year = {2023}
}

@inproceedings{yang2024cogvideox,
  title = {{CogVideoX: Text-to-Video Diffusion Models with An Expert Transformer}},
  author = {Yang, Zhuoyi and Teng, Jiayan and Zheng, Wendi and Ding, Ming and Huang, Shiyu and Xu, Jiazheng and Yang, Yuanming and Hong, Wenyi and Zhang, Xiaohan and Feng, Guanyu and Yin, Da and Zhang, Yuxuan and Wang, Weihan and Cheng, Yean and Xu, Bin and Gu, Xiaotao and Dong, Yuxiao and Tang, Jie},
  booktitle = ICLR,
  year = {2025}
}

@article{lin2024open,
  title = {{Open-Sora Plan: Open-Source Large Video Generation Model}},
  author = {Lin, Bin and Ge, Yunyang and Cheng, Xinhua and Li, Zongjian and Zhu, Bin and Wang, Shaodong and He, Xianyi and Ye, Yang and Yuan, Shenghai and Chen, Liuhan and Jia, Tanghui and Zhang, Junwu and Tang, Zhenyu and Pang, Yatian and She, Bin and Yan, Cen and Hu, Zhiheng and Dong, Xiaoyi and Chen, Lin and Pan, Zhang and Zhou, Xing and Dong, Shaoling and Tian, Yonghong and Yuan, Li},
  journal = {arXiv:2412.00131},
  year = {2024}
}

@inproceedings{gao2024cat3d,
  title = {{CAT3D: Create Anything in 3D with Multi-View Diffusion Models}},
  author = {Gao, Ruiqi and Ho{\l}y{\'n}ski, Aleksander and Henzler, Philipp and Brussee, Arthur and Martin-Brualla, Ricardo and Srinivasan, Pratul and Barron, Jonathan T. and Poole, Ben},
  booktitle = NIPS,
  year = {2024}
}

@inproceedings{zeronvs,
  title = {{ZeroNVS: Zero-Shot 360-Degree View Synthesis from a Single Image}},
  author = {Sargent, Kyle and Li, Zizhang and Shah, Tanmay and Herrmann, Charles and Yu, Hong-Xing and Zhang, Yunzhi and Chan, Eric Ryan and Lagun, Dmitry and Fei-Fei, Li and Sun, Deqing and Wu, Jiajun},
  booktitle = CVPR,
  year = {2024}
}

@inproceedings{seva,
  title = {{Stable Virtual Camera: Generative View Synthesis with Diffusion Models}},
  author = {Zhou, Jensen and Gao, Hang and Voleti, Vikram and Vasishta, Aaryaman and Yao, Chun-Han and Boss, Mark and Torr, Philip and Rupprecht, Christian and Jampani, Varun},
  booktitle = ICCV,
  year = {2025}
}

@inproceedings{wang2023motionctrl,
  title = {{MotionCtrl: A Unified and Flexible Motion Controller for Video Generation}},
  author = {Wang, Zhouxia and Yuan, Ziyang and Wang, Xintao and Li, Yaowei and Chen, Tianshui and Xia, Menghan and Luo, Ping and Shan, Ying},
  booktitle = SIGGRAPH,
  year = {2024}
}

@inproceedings{wu2024reconfusion,
  title = {{ReconFusion: 3D Reconstruction with Diffusion Priors}},
  author = {Wu, Rundi and Mildenhall, Ben and Henzler, Philipp and Park, Keunhong and Gao, Ruiqi and Watson, Daniel and Srinivasan, Pratul P. and Verbin, Dor and Barron, Jonathan T. and Poole, Ben and Ho{\l}y{\'n}ski, Aleksander},
  booktitle = CVPR,
  year = {2024}
}

@inproceedings{sun2024dimensionx,
  title = {{DimensionX: Create Any 3D and 4D Scenes from a Single Image with Decoupled Video Diffusion}},
  author = {Sun, Wenqiang and Chen, Shuo and Liu, Fangfu and Chen, Zilong and Duan, Yueqi and Zhu, Jun and Zhang, Jun and Wang, Yikai},
  booktitle = ICCV,
  year = {2025}
}

@inproceedings{bai2025recammaster,
  title = {{ReCamMaster: Camera-Controlled Generative Rendering from A Single Video}},
  author = {Bai, Jianhong and Xia, Menghan and Fu, Xiao and Wang, Xintao and Mu, Lianrui and Cao, Jinwen and Liu, Zuozhu and Hu, Haoji and Bai, Xiang and Wan, Pengfei and Zhang, Di},
  booktitle = ICCV,
  year = {2025}
}

@inproceedings{gcd,
  title = {{Generative Camera Dolly: Extreme Monocular Dynamic Novel View Synthesis}},
  author = {Van Hoorick, Basile and Wu, Rundi and Ozguroglu, Ege and Sargent, Kyle and Liu, Ruoshi and Tokmakov, Pavel and Dave, Achal and Zheng, Changxi and Vondrick, Carl},
  booktitle = ECCV,
  year = {2024}
}

@inproceedings{wang2024dust3r,
  title = {{DUSt3R: Geometric 3D Vision Made Easy}},
  author = {Wang, Shuzhe and Leroy, Vincent and Cabon, Yohann and Chidlovskii, Boris and Revaud, Jerome},
  booktitle = CVPR,
  year = {2024}
}

@inproceedings{hu2024depthcrafter,
  title = {{DepthCrafter: Generating Consistent Long Depth Sequences for Open-world Videos}},
  author = {Hu, Wenbo and Gao, Xiangjun and Li, Xiaoyu and Zhao, Sijie and Cun, Xiaodong and Zhang, Yong and Quan, Long and Shan, Ying},
  booktitle = CVPR,
  year = {2025}
}

@inproceedings{dptv2,
  title = {{Depth Anything V2}},
  author = {Yang, Lihe and Kang, Bingyi and Huang, Zilong and Zhao, Zhen and Xu, Xiaogang and Feng, Jiashi and Zhao, Hengshuang},
  booktitle = NIPS,
  year = {2024}
}

@inproceedings{wang2025vggt,
  title = {{VGGT: Visual Geometry Grounded Transformer}},
  author = {Wang, Jianyuan and Chen, Minghao and Karaev, Nikita and Vedaldi, Andrea and Rupprecht, Christian and Novotny, David},
  booktitle = CVPR,
  year = {2025}
}

@article{huang2025vipe,
  title = {{ViPE: Video Pose Engine for 3D Geometric Perception}},
  author = {Huang, Jiahui and Zhou, Qunjie and Rabeti, Hesam and Korovko, Aleksandr and Ling, Huan and Ren, Xuanchi and Shen, Tianchang and Gao, Jun and Slepichev, Dmitry and Lin, Chen-Hsuan and Ren, Jiawei and Xie, Kevin and Biswas, Joydeep and Leal-Taixe, Laura and Fidler, Sanja},
  journal = {arXiv:2508.10934},
  year = {2025}
}

@inproceedings{lan2025stream3r,
  title = {{STream3R: Scalable Sequential 3D Reconstruction with Causal Transformer}},
  author = {Lan, Yushi and Luo, Yihang and Hong, Fangzhou and Zhou, Shangchen and Chen, Honghua and Lyu, Zhaoyang and Yang, Shuai and Dai, Bo and Loy, Chen Change and Pan, Xingang},
  booktitle = ICLR,
  year = {2026}
}

@inproceedings{wang2025moge,
  title = {{MoGe: Unlocking Accurate Monocular Geometry Estimation for Open-Domain Images with Optimal Training Supervision}},
  author = {Wang, Ruicheng and Xu, Sicheng and Dai, Cassie and Xiang, Jianfeng and Deng, Yu and Tong, Xin and Yang, Jiaolong},
  booktitle = CVPR,
  year = {2025}
}

@article{yu2024viewcrafter,
  title = {{ViewCrafter: Taming Video Diffusion Models for High-fidelity Novel View Synthesis}},
  author = {Yu, Wangbo and Xing, Jinbo and Yuan, Li and Hu, Wenbo and Li, Xiaoyu and Huang, Zhipeng and Gao, Xiangjun and Wong, Tien-Tsin and Shan, Ying and Tian, Yonghong},
  journal = PAMI,
  year = {2025}
}

@inproceedings{ren2025gen3c,
  title = {{GEN3C: 3D-Informed World-Consistent Video Generation with Precise Camera Control}},
  author = {Ren, Xuanchi and Shen, Tianchang and Huang, Jiahui and Ling, Huan and Lu, Yifan and Nimier-David, Merlin and M\"uller, Thomas and Keller, Alexander and Fidler, Sanja and Gao, Jun},
  booktitle = CVPR,
  year = {2025}
}

@inproceedings{uni3c,
  title = {{Uni3C: Unifying Precisely 3D-Enhanced Camera and Human Motion Controls for Video Generation}},
  author = {Cao, Chenjie and Zhou, Jingkai and Li, Shikai and Liang, Jingyun and Yu, Chaohui and Wang, Fan and Xue, Xiangyang and Fu, Yanwei},
  booktitle = SIGGRAPHASIA,
  year = {2025}
}

@inproceedings{you2024solver,
  title = {{NVS-Solver: Video Diffusion Model as Zero-Shot Novel View Synthesizer}},
  author = {You, Meng and Zhu, Zhiyu and Liu, Hui and Hou, Junhui},
  booktitle = ICLR,
  year = {2025}
}

@inproceedings{muller2024multidiff,
  title = {{MultiDiff: Consistent Novel View Synthesis from a Single Image}},
  author = {M\"uller, Norman and Schwarz, Katja and R\"ossle, Barbara and Porzi, Lorenzo and Bul\`o, Samuel Rota and Nie{\ss}ner, Matthias and Kontschieder, Peter},
  booktitle = CVPR,
  year = {2024}
}

@inproceedings{Ma2024See3D,
  title = {{You See it, You Got it: Learning 3D Creation on Pose-Free Videos at Scale}},
  author = {Ma, Baorui and Gao, Huachen and Deng, Haoge and Luo, Zhengxiong and Huang, Tiejun and Tang, Lulu and Wang, Xinlong},
  booktitle = CVPR,
  year = {2025}
}

@inproceedings{lei2024mosca,
  title = {{MoSca: Dynamic Gaussian Fusion from Casual Videos via 4D Motion Scaffolds}},
  author = {Lei, Jiahui and Weng, Yijia and Harley, Adam W. and Guibas, Leonidas and Daniilidis, Kostas},
  booktitle = CVPR,
  year = {2025}
}

@inproceedings{mark2025trajectorycrafter,
  title = {{TrajectoryCrafter: Redirecting Camera Trajectory for Monocular Videos via Diffusion Models}},
  author = {Yu, Mark and Hu, Wenbo and Xing, Jinbo and Shan, Ying},
  booktitle = ICCV,
  year = {2025}
}

@inproceedings{vace,
  title = {{VACE: All-in-One Video Creation and Editing}},
  author = {Jiang, Zeyinzi and Han, Zhen and Mao, Chaojie and Zhang, Jingfeng and Pan, Yulin and Liu, Yu},
  booktitle = ICCV,
  year = {2025}
}

@article{wan2025,
  title = {{Wan: Open and Advanced Large-Scale Video Generative Models}},
  author = {{Team Wan} and Wang, Ang and Ai, Baole and Wen, Bin and Mao, Chaojie and Xie, Chen-Wei and Chen, Di and Yu, Feiwu and Zhao, Haiming and Yang, Jianxiao and Zeng, Jianyuan and Wang, Jiayu and Zhang, Jingfeng and Zhou, Jingren and Wang, Jinkai and Chen, Jixuan and Zhu, Kai and Zhao, Kang and Yan, Keyu and Huang, Lianghua and Feng, Mengyang and Zhang, Ningyi and Li, Pandeng and Wu, Pingyu and Chu, Ruihang and Feng, Ruili and Zhang, Shiwei and Sun, Siyang and Fang, Tao and Wang, Tianxing and Gui, Tianyi and Weng, Tingyu and Shen, Tong and Lin, Wei and Wang, Wei and Wang, Wei and Zhou, Wenmeng and Wang, Wente and Shen, Wenting and Yu, Wenyuan and Shi, Xianzhong and Huang, Xiaoming and Xu, Xin and Kou, Yan and Lv, Yangyu and Li, Yifei and Liu, Yijing and Wang, Yiming and Zhang, Yingya and Huang, Yitong and Li, Yong and Wu, You and Liu, Yu and Pan, Yulin and Zheng, Yun and Hong, Yuntao and Shi, Yupeng and Feng, Yutong and Jiang, Zeyinzi and Han, Zhen and Wu, Zhi-Fan and Liu, Ziyu},
  journal = {arXiv:2503.20314},
  year = {2025}
}

@inproceedings{barron2021mip,
  title = {{Mip-NeRF: A Multiscale Representation for Anti-Aliasing Neural Radiance Fields}},
  author = {Barron, Jonathan T. and Mildenhall, Ben and Tancik, Matthew and Hedman, Peter and Martin-Brualla, Ricardo and Srinivasan, Pratul P.},
  booktitle = ICCV,
  year = {2021}
}

@inproceedings{barron2022mip,
  title = {{Mip-NeRF 360: Unbounded Anti-Aliased Neural Radiance Fields}},
  author = {Barron, Jonathan T. and Mildenhall, Ben and Verbin, Dor and Srinivasan, Pratul P. and Hedman, Peter},
  booktitle = CVPR,
  year = {2022}
}

@inproceedings{verbin2022ref,
  title = {{Ref-NeRF: Structured View-Dependent Appearance for Neural Radiance Fields}},
  author = {Verbin, Dor and Hedman, Peter and Mildenhall, Ben and Zickler, Todd and Barron, Jonathan T. and Srinivasan, Pratul P.},
  booktitle = CVPR,
  year = {2022}
}

@inproceedings{barron2023zip,
  title = {{Zip-NeRF: Anti-Aliased Grid-Based Neural Radiance Fields}},
  author = {Barron, Jonathan T. and Mildenhall, Ben and Verbin, Dor and Srinivasan, Pratul P. and Hedman, Peter},
  booktitle = ICCV,
  year = {2023}
}

@inproceedings{hu2023Tri-MipRF,
  title = {{Tri-MipRF: Tri-Mip Representation for Efficient Anti-Aliasing Neural Radiance Fields}},
  author = {Hu, Wenbo and Wang, Yuling and Ma, Lin and Yang, Bangbang and Gao, Lin and Liu, Xiao and Ma, Yuewen},
  booktitle = ICCV,
  year = {2023}
}

@inproceedings{liu2024ripnerf,
  title = {{Rip-NeRF: Anti-aliasing Radiance Fields with Ripmap-Encoded Platonic Solids}},
  author = {Liu, Junchen and Hu, Wenbo and Yang, Zhuo and Chen, Jianteng and Wang, Guoliang and Chen, Xiaoxue and Cai, Yantong and Gao, Huan-ang and Zhao, Hao},
  booktitle = SIGGRAPH,
  year = {2024}
}

@inproceedings{liang2025analytic,
  title = {{Analytic-Splatting: Anti-Aliased 3D Gaussian Splatting via Analytic Integration}},
  author = {Liang, Zhihao and Zhang, Qi and Hu, Wenbo and Zhu, Lei and Feng, Ying and Jia, Kui},
  booktitle = ECCV,
  year = {2024}
}

@inproceedings{zhang2024pixelgs,
  title = {{Pixel-GS: Density Control with Pixel-Aware Gradient for 3D Gaussian Splatting}},
  author = {Zhang, Zheng and Hu, Wenbo and Lao, Yixing and He, Tong and Zhao, Hengshuang},
  booktitle = ECCV,
  year = {2024}
}

@inproceedings{yu2024mip,
  title = {{Mip-Splatting: Alias-free 3D Gaussian Splatting}},
  author = {Yu, Zehao and Chen, Anpei and Huang, Binbin and Sattler, Torsten and Geiger, Andreas},
  booktitle = CVPR,
  year = {2024}
}

@article{muller2022instant,
  title = {{Instant neural graphics primitives with a multiresolution hash encoding}},
  author = {M{\"u}ller, Thomas and Evans, Alex and Schied, Christoph and Keller, Alexander},
  journal = TOG,
  year = {2022}
}

@article{fan2024instantsplat,
  title = {{InstantSplat: Sparse-view Gaussian Splatting in Seconds}},
  author = {Fan, Zhiwen and Cong, Wenyan and Wen, Kairun and Wang, Kevin and Zhang, Jian and Ding, Xinghao and Xu, Danfei and Ivanovic, Boris and Pavone, Marco and Pavlakos, Georgios and Wang, Zhangyang and Wang, Yue},
  journal = {arXiv:2403.20309},
  year = {2024}
}

@inproceedings{yu2023nofa,
  title = {{NOFA: NeRF-based One-shot Facial Avatar Reconstruction}},
  author = {Yu, Wangbo and Fan, Yanbo and Zhang, Yong and Wang, Xuan and Yin, Fei and Bai, Yunpeng and Cao, Yan-Pei and Shan, Ying and Wu, Yang and Sun, Zhongqian and Wu, Baoyuan},
  booktitle = SIGGRAPH,
  year = {2023}
}

@inproceedings{yu2024evagaussians,
  title = {{EvaGaussians: Event Stream Assisted Gaussian Splatting from Blurry Images}},
  author = {Yu, Wangbo and Feng, Chaoran and Li, Jianing and Tang, Jiye and Yang, Jiashu and Tang, Zhenyu and Cao, Meng and Jia, Xu and Yang, Yuchao and Yuan, Li and Tian, Yonghong},
  booktitle = ICCV,
  year = {2025}
}

@inproceedings{feng2025ae,
  title = {{AE-NeRF: Augmenting Event-Based Neural Radiance Fields for Non-ideal Conditions and Larger Scenes}},
  author = {Feng, Chaoran and Yu, Wangbo and Cheng, Xinhua and Tang, Zhenyu and Zhang, Junwu and Yuan, Li and Tian, Yonghong},
  booktitle = AAAI,
  year = {2025}
}

@article{li2022nerfacc,
  title = {{NerfAcc: A General NeRF Acceleration Toolbox}},
  author = {Li, Ruilong and Tancik, Matthew and Kanazawa, Angjoo},
  journal = {arXiv:2210.04847},
  year = {2022}
}

@inproceedings{chen2021mvsnerf,
  title = {{MVSNeRF: Fast Generalizable Radiance Field Reconstruction From Multi-View Stereo}},
  author = {Chen, Anpei and Xu, Zexiang and Zhao, Fuqiang and Zhang, Xiaoshuai and Xiang, Fanbo and Yu, Jingyi and Su, Hao},
  booktitle = ICCV,
  year = {2021}
}

@inproceedings{wang2023sparsenerf,
  title = {{SparseNeRF: Distilling Depth Ranking for Few-shot Novel View Synthesis}},
  author = {Wang, Guangcong and Chen, Zhaoxi and Loy, Chen Change and Liu, Ziwei},
  booktitle = ICCV,
  year = {2023}
}

@inproceedings{zhu2023fsgs,
  title = {{FSGS: Real-Time Few-Shot View Synthesis Using Gaussian Splatting}},
  author = {Zhu, Zehao and Fan, Zhiwen and Jiang, Yifan and Wang, Zhangyang},
  booktitle = ECCV,
  year = {2024}
}

@inproceedings{lin2021barf,
  title = {{BARF: Bundle-Adjusting Neural Radiance Fields}},
  author = {Lin, Chen-Hsuan and Ma, Wei-Chiu and Torralba, Antonio and Lucey, Simon},
  booktitle = ICCV,
  year = {2021}
}

@inproceedings{garbin2021fastnerf,
  title = {{FastNeRF: High-Fidelity Neural Rendering at 200FPS}},
  author = {Garbin, Stephan J. and Kowalski, Marek and Johnson, Matthew and Shotton, Jamie and Valentin, Julien},
  booktitle = ICCV,
  year = {2021}
}

@inproceedings{chen2024mvsplat,
  title = {{MVSplat: Efficient 3D Gaussian Splatting from Sparse Multi-view Images}},
  author = {Chen, Yuedong and Xu, Haofei and Zheng, Chuanxia and Zhuang, Bohan and Pollefeys, Marc and Geiger, Andreas and Cham, Tat-Jen and Cai, Jianfei},
  booktitle = ECCV,
  year = {2024}
}

@inproceedings{hyperreel,
  title = {{HyperReel: High-Fidelity 6-DoF Video With Ray-Conditioned Sampling}},
  author = {Attal, Benjamin and Huang, Jia-Bin and Richardt, Christian and Zollh\"ofer, Michael and Kopf, Johannes and O{\textquoteright}Toole, Matthew and Kim, Changil},
  booktitle = CVPR,
  year = {2023}
}

@inproceedings{fridovich2023kplane,
  title = {{K-Planes: Explicit Radiance Fields in Space, Time, and Appearance}},
  author = {Fridovich-Keil, Sara and Meanti, Giacomo and Warburg, Frederik Rahb{\ae}k and Recht, Benjamin and Kanazawa, Angjoo},
  booktitle = CVPR,
  year = {2023}
}

@inproceedings{cao2023hexplane,
  title = {{HexPlane: A Fast Representation for Dynamic Scenes}},
  author = {Cao, Ang and Johnson, Justin},
  booktitle = CVPR,
  year = {2023}
}

@inproceedings{li2024spacetime,
  title = {{Spacetime Gaussian Feature Splatting for Real-Time Dynamic View Synthesis}},
  author = {Li, Zhan and Chen, Zhang and Li, Zhong and Xu, Yi},
  booktitle = CVPR,
  year = {2024}
}

@inproceedings{pumarola2021d,
  title = {{D-NeRF: Neural Radiance Fields for Dynamic Scenes}},
  author = {Pumarola, Albert and Corona, Enric and Pons-Moll, Gerard and Moreno-Noguer, Francesc},
  booktitle = CVPR,
  year = {2021}
}

@article{nerfplayer,
  title = {{NeRFPlayer: A Streamable Dynamic Scene Representation with Decomposed Neural Radiance Fields}},
  author = {Song, Liangchen and Chen, Anpei and Li, Zhong and Chen, Zhang and Chen, Lele and Yuan, Junsong and Xu, Yi and Geiger, Andreas},
  journal = TVCG,
  year = {2023}
}

@inproceedings{yang2023gs4d,
  title = {{Real-time Photorealistic Dynamic Scene Representation and Rendering with 4D Gaussian Splatting}},
  author = {Yang, Zeyu and Yang, Hongye and Pan, Zijie and Zhang, Li},
  booktitle = ICLR,
  year = {2024}
}

@inproceedings{yoon2020nvidia,
  title = {{Novel View Synthesis of Dynamic Scenes With Globally Coherent Depths From a Monocular Camera}},
  author = {Shin Yoon, Jae and Kim, Kihwan and Gallo, Orazio and Park, Hyun Soo and Kautz, Jan},
  booktitle = CVPR,
  year = {2020}
}

@inproceedings{sceneflow,
  title = {{Neural Scene Flow Fields for Space-Time View Synthesis of Dynamic Scenes}},
  author = {Li, Zhengqi and Niklaus, Simon and Snavely, Noah and Wang, Oliver},
  booktitle = CVPR,
  year = {2021}
}

@inproceedings{nerfies,
  title = {{Nerfies: Deformable Neural Radiance Fields}},
  author = {Park, Keunhong and Sinha, Utkarsh and Barron, Jonathan T. and Bouaziz, Sofien and Goldman, Dan B and Seitz, Steven M. and Martin-Brualla, Ricardo},
  booktitle = ICCV,
  year = {2021}
}

@inproceedings{Tretschk_2021_ICCV,
  title = {{Non-Rigid Neural Radiance Fields: Reconstruction and Novel View Synthesis of a Dynamic Scene From Monocular Video}},
  author = {Tretschk, Edgar and Tewari, Ayush and Golyanik, Vladislav and Zollh\"ofer, Michael and Lassner, Christoph and Theobalt, Christian},
  booktitle = ICCV,
  year = {2021}
}

@inproceedings{DynIBaR,
  title = {{DynIBaR: Neural Dynamic Image-Based Rendering}},
  author = {Li, Zhengqi and Wang, Qianqian and Cole, Forrester and Tucker, Richard and Snavely, Noah},
  booktitle = CVPR,
  year = {2023}
}

@inproceedings{lee2025fast,
  title = {{Fast View Synthesis of Casual Videos with Soup-of-Planes}},
  author = {Lee, Yao-Chih and Zhang, Zhoutong and Blackburn-Matzen, Kevin and Niklaus, Simon and Zhang, Jianming and Huang, Jia-Bin and Liu, Feng},
  booktitle = ECCV,
  year = {2024}
}

@inproceedings{stearns2024marbles,
  title = {{Dynamic Gaussian Marbles for Novel View Synthesis of Casual Monocular Videos}},
  author = {Stearns, Colton and Harley, Adam W. and Uy, Mikaela and Dubost, Florian and Tombari, Federico and Wetzstein, Gordon and Guibas, Leonidas},
  booktitle = SIGGRAPHASIA,
  year = {2024}
}

@article{gao2024gaussianflow,
  title = {{GaussianFlow: Splatting Gaussian Dynamics for 4D Content Creation}},
  author = {Gao, Quankai and Xu, Qiangeng and Cao, Zhe and Mildenhall, Ben and Ma, Wenchao and Chen, Le and Tang, Danhang and Neumann, Ulrich},
  journal = {arXiv:2403.12365},
  year = {2024}
}

@article{xu20254dgt,
  title = {{4DGT: Learning a 4D Gaussian Transformer Using Real-World Monocular Videos}},
  author = {Xu, Zhen and Li, Zhengqin and Dong, Zhao and Zhou, Xiaowei and Newcombe, Richard and Lv, Zhaoyang},
  journal = {arXiv:2506.08015},
  year = {2025}
}

@inproceedings{huang2025vivid4d,
  title = {{Vivid4D: Improving 4D Reconstruction from Monocular Video by Video Inpainting}},
  author = {Huang, Jiaxin and Miao, Sheng and Yang, Bangbang and Ma, Yuewen and Liao, Yiyi},
  booktitle = ICCV,
  year = {2025}
}

@inproceedings{nazarczuk2025vidar,
  title = {{ViDAR: Video Diffusion-Aware 4D Reconstruction From Monocular Inputs}},
  author = {Nazarczuk, Michal and Catley-Chandar, Sibi and Tanay, Thomas and Zhang, Zhensong and Slabaugh, Greg and P{\'e}rez-Pellitero, Eduardo},
  booktitle = NIPS,
  year = {2025}
}

@inproceedings{yang2026neoverse,
  title = {{NeoVerse: Enhancing 4D World Model with in-the-wild Monocular Videos}},
  author = {Yang, Yuxue and Fan, Lue and Shi, Ziqi and Peng, Junran and Wang, Feng and Zhang, Zhaoxiang},
  booktitle = CVPR,
  year = {2026}
}

@inproceedings{zhu2026worldfrommotion,
  title = {{World from Motion: Generative Dynamic Gaussian Reconstruction from Monocular Video}},
  author = {Zhu, Liyuan and Huang, Shengyu and Mazumdar, Amrita and Li, Tianye and Gojcic, Zan and Wetzstein, Gordon and Armeni, Iro and De Mello, Shalini and Trevithick, Alex},
  booktitle = NIPS,
  year = {2026}
}

@inproceedings{wiles2020synsin,
  title = {{SynSin: End-to-End View Synthesis From a Single Image}},
  author = {Wiles, Olivia and Gkioxari, Georgia and Szeliski, Richard and Johnson, Justin},
  booktitle = CVPR,
  year = {2020}
}

@inproceedings{rombach2021geometryfree,
  title = {{Geometry-Free View Synthesis: Transformers and No 3D Priors}},
  author = {Rombach, Robin and Esser, Patrick and Ommer, Bj\"orn},
  booktitle = ICCV,
  year = {2021}
}

@inproceedings{rockwell2021pixelsynth,
  title = {{PixelSynth: Generating a 3D-Consistent Experience From a Single Image}},
  author = {Rockwell, Chris and Fouhey, David F. and Johnson, Justin},
  booktitle = ICCV,
  year = {2021}
}

@article{park2024bridging,
  title = {{Bridging Implicit and Explicit Geometric Transformation for Single-Image View Synthesis}},
  author = {Park, Byeongjun and Go, Hyojun and Kim, Changick},
  journal = PAMI,
  year = {2024}
}

@inproceedings{ho2020denoising,
  title = {{Denoising Diffusion Probabilistic Models}},
  author = {Ho, Jonathan and Jain, Ajay and Abbeel, Pieter},
  booktitle = NIPS,
  year = {2020}
}

@inproceedings{song2021denoising,
  title = {{Denoising Diffusion Implicit Models}},
  author = {Song, Jiaming and Meng, Chenlin and Ermon, Stefano},
  booktitle = ICLR,
  year = {2021}
}

@inproceedings{huang2024roompainter,
  title = {{RoomPainter: View-Integrated Diffusion for Consistent Indoor Scene Texturing}},
  author = {Huang, Zhipeng and Yu, Wangbo and Cheng, Xinhua and Zhao, Chengshu and Ge, Yunyang and Guo, Mingyi and Yuan, Li and Tian, Yonghong},
  booktitle = CVPR,
  year = {2025}
}

@article{zhou2024holodreamer,
  title = {{HoloDreamer: Holistic 3D Panoramic World Generation from Text Descriptions}},
  author = {Zhou, Haiyang and Cheng, Xinhua and Yu, Wangbo and Tian, Yonghong and Yuan, Li},
  journal = {arXiv:2407.15187},
  year = {2024}
}

@inproceedings{yu2024hifi,
  title = {{HiFi-123: Towards High-Fidelity One Image to 3D Content Generation}},
  author = {Yu, Wangbo and Yuan, Li and Cao, Yan-Pei and Gao, Xiangjun and Li, Xiaoyu and Hu, Wenbo and Quan, Long and Shan, Ying and Tian, Yonghong},
  booktitle = ECCV,
  year = {2024}
}

@inproceedings{liu2023zero,
  title = {{Zero-1-to-3: Zero-shot One Image to 3D Object}},
  author = {Liu, Ruoshi and Wu, Rundi and Van Hoorick, Basile and Tokmakov, Pavel and Zakharov, Sergey and Vondrick, Carl},
  booktitle = ICCV,
  year = {2023}
}

@inproceedings{xing2023dynamicrafter,
  title = {{DynamiCrafter: Animating Open-domain Images with Video Diffusion Priors}},
  author = {Xing, Jinbo and Xia, Menghan and Zhang, Yong and Chen, Haoxin and Yu, Wangbo and Liu, Hanyuan and Liu, Gongye and Wang, Xintao and Shan, Ying and Wong, Tien-Tsin},
  booktitle = ECCV,
  year = {2024}
}

@inproceedings{duan2025worldscore,
  title = {{WorldScore: A Unified Evaluation Benchmark for World Generation}},
  author = {Duan, Haoyi and Yu, Hong-Xing and Chen, Sirui and Fei-Fei, Li and Wu, Jiajun},
  booktitle = ICCV,
  year = {2025}
}

@misc{cogfun,
  title = {{CogVideoX-Fun}},
  author = {{aigc-apps}},
  howpublished = {\url{https://github.com/aigc-apps/CogVideoX-Fun}},
  year = {2024}
}

@inproceedings{bahmani2024vd3d,
  title = {{VD3D: Taming Large Video Diffusion Transformers for 3D Camera Control}},
  author = {Bahmani, Sherwin and Skorokhodov, Ivan and Siarohin, Aliaksandr and Menapace, Willi and Qian, Guocheng and Vasilkovsky, Michael and Lee, Hsin-Ying and Wang, Chaoyang and Zou, Jiaxu and Tagliasacchi, Andrea and Lindell, David and Tulyakov, Sergey},
  booktitle = ICLR,
  year = {2025}
}

@article{xu2024camco,
  title = {{CamCo: Camera-Controllable 3D-Consistent Image-to-Video Generation}},
  author = {Xu, Dejia and Nie, Weili and Liu, Chao and Liu, Sifei and Kautz, Jan and Wang, Zhangyang and Vahdat, Arash},
  journal = {arXiv:2406.02509},
  year = {2024}
}

@inproceedings{bai2024syncammaster,
  title = {{SynCamMaster: Synchronizing Multi-Camera Video Generation from Diverse Viewpoints}},
  author = {Bai, Jianhong and Xia, Menghan and Wang, Xintao and Yuan, Ziyang and Liu, Zuozhu and Hu, Haoji and Wan, Pengfei and Zhang, Di},
  booktitle = ICLR,
  year = {2025}
}

@inproceedings{he2024cameractrl,
  title = {{CameraCtrl: Enabling Camera Control for Video Diffusion Models}},
  author = {He, Hao and Xu, Yinghao and Guo, Yuwei and Wetzstein, Gordon and Dai, Bo and Li, Hongsheng and Yang, Ceyuan},
  booktitle = ICLR,
  year = {2025}
}

@inproceedings{liang2024wonderland,
  title = {{Wonderland: Navigating 3D Scenes from a Single Image}},
  author = {Liang, Hanwen and Cao, Junli and Goel, Vidit and Qian, Guocheng and Korolev, Sergei and Terzopoulos, Demetri and Plataniotis, Konstantinos N. and Tulyakov, Sergey and Ren, Jian},
  booktitle = CVPR,
  year = {2025}
}

@inproceedings{zhao2024genxd,
  title = {{GenXD: Generating Any 3D and 4D Scenes}},
  author = {Zhao, Yuyang and Lin, Chung-Ching and Lin, Kevin and Yan, Zhiwen and Li, Linjie and Yang, Zhengyuan and Wang, Jianfeng and Lee, Gim Hee and Wang, Lijuan},
  booktitle = ICLR,
  year = {2025}
}

@inproceedings{wu2024cat4d,
  title = {{CAT4D: Create Anything in 4D with Multi-View Video Diffusion Models}},
  author = {Wu, Rundi and Gao, Ruiqi and Poole, Ben and Trevithick, Alex and Zheng, Changxi and Barron, Jonathan T. and Holynski, Aleksander},
  booktitle = CVPR,
  year = {2025}
}

@inproceedings{li2026cameraanything,
  title = {{CameraAnything: Refilming Videos with Arbitrary Camera Control}},
  author = {Li, Yixuan and Zeng, Yanhong and Cheng, Ka Leong and Zhu, Jiayi and Wang, Hanlin and Wang, Wen and Meng, Yihao and Ouyang, Hao and Wang, Qiuyu and Yu, Yue and Wang, ZiDong and Zhang, Yiyuan and Shen, Yujun and Lin, Dahua},
  booktitle = ECCV,
  year = {2026}
}

@inproceedings{huang2026spacetimepilot,
  title = {{SpaceTimePilot: Generative Rendering of Dynamic Scenes Across Space and Time}},
  author = {Huang, Zhening and Jeong, Hyeonho and Chen, Xuelin and Gryaditskaya, Yulia and Wang, Tuanfeng Y. and Lasenby, Joan and Huang, Chun-Hao},
  booktitle = CVPR,
  year = {2026}
}

@inproceedings{gu2025das,
  title = {{Diffusion as Shader: 3D-aware Video Diffusion for Versatile Video Generation Control}},
  author = {Gu, Zekai and Yan, Rui and Lu, Jiahao and Li, Peng and Dou, Zhiyang and Si, Chenyang and Dong, Zhen and Liu, Qifeng and Lin, Cheng and Liu, Ziwei and Wang, Wenping and Liu, Yuan},
  booktitle = SIGGRAPH,
  year = {2025}
}

@inproceedings{xiao2024trajectory,
  title = {{Trajectory attention for fine-grained video motion control}},
  author = {Xiao, Zeqi and Ouyang, Wenqi and Zhou, Yifan and Yang, Shuai and Yang, Lei and Si, Jianlou and Pan, Xingang},
  booktitle = ICLR,
  year = {2025}
}

@inproceedings{wang2025videoscene,
  title = {{VideoScene: Distilling Video Diffusion Model to Generate 3D Scenes in One Step}},
  author = {Wang, Hanyang and Liu, Fangfu and Chi, Jiawei and Duan, Yueqi},
  booktitle = CVPR,
  year = {2025}
}

@inproceedings{xie2026lavr,
  title = {{LaVR: Scene Latent Conditioned Generative Video Trajectory Re-Rendering using Large 4D Reconstruction Models}},
  author = {Xie, Mingyang and Khan, Numair and Wang, Tianfu and Dhingra, Naina and Nam, Seonghyeon and Yang, Haitao and Hui, Zhuo and Metzler, Christopher and Vedaldi, Andrea and Pirsiavash, Hamed and Luo, Lei},
  booktitle = CVPR,
  year = {2026}
}

@inproceedings{yu2021pixelnerf,
  title = {{pixelNeRF: Neural Radiance Fields From One or Few Images}},
  author = {Yu, Alex and Ye, Vickie and Tancik, Matthew and Kanazawa, Angjoo},
  booktitle = CVPR,
  year = {2021}
}

@inproceedings{zhang2024recapture,
  title = {{ReCapture: Generative Video Camera Controls for User-Provided Videos using Masked Video Fine-Tuning}},
  author = {Zhang, David Junhao and Paiss, Roni and Zada, Shiran and Karnad, Nikhil and Jacobs, David E. and Pritch, Yael and Mosseri, Inbar and Shou, Mike Zheng and Wadhwa, Neal and Ruiz, Nataniel},
  booktitle = CVPR,
  year = {2025}
}

@inproceedings{yu2025wonderworld,
  title = {{WonderWorld: Interactive 3D Scene Generation from a Single Image}},
  author = {Yu, Hong-Xing and Duan, Haoyi and Herrmann, Charles and Freeman, William T. and Wu, Jiajun},
  booktitle = CVPR,
  year = {2025}
}

@inproceedings{wang2025vistadream,
  title = {{VistaDream: Sampling multiview consistent images for single-view scene reconstruction}},
  author = {Wang, Haiping and Liu, Yuan and Liu, Ziwei and Wang, Wenping and Dong, Zhen and Yang, Bisheng},
  booktitle = ICCV,
  year = {2025}
}

@article{shriram2024realmdreamer,
  title = {{RealmDreamer: Text-Driven 3D Scene Generation with Inpainting and Depth Diffusion}},
  author = {Shriram, Jaidev and Trevithick, Alex and Liu, Lingjie and Ramamoorthi, Ravi},
  journal = {arXiv:2404.07199},
  year = {2024}
}

@article{chung2023luciddreamer,
  title = {{LucidDreamer: Domain-Free Generation of 3D Gaussian Splatting Scenes}},
  author = {Chung, Jaeyoung and Lee, Suyoung and Nam, Hyeongjin and Lee, Jaerin and Lee, Kyoung Mu},
  journal = TVCG,
  year = {2025}
}

@inproceedings{liu2025free4d,
  title = {{Free4D: Tuning-free 4D Scene Generation with Spatial-Temporal Consistency}},
  author = {Liu, Tianqi and Huang, Zihao and Chen, Zhaoxi and Wang, Guangcong and Hu, Shoukang and Shen, Liao and Sun, Huiqiang and Cao, Zhiguo and Li, Wei and Liu, Ziwei},
  booktitle = ICCV,
  year = {2025}
}

@inproceedings{tung2024megascenes,
  title = {{MegaScenes: Scene-Level View Synthesis at Scale}},
  author = {Tung, Joseph and Chou, Gene and Cai, Ruojin and Yang, Guandao and Zhang, Kai and Wetzstein, Gordon and Hariharan, Bharath and Snavely, Noah},
  booktitle = ECCV,
  year = {2024}
}

@inproceedings{zhang2025spatialcrafter,
  title = {{SpatialCrafter: Unleashing the Imagination of Video Diffusion Models for Scene Reconstruction from Limited Observations}},
  author = {Zhang, Songchun and Xu, Huiyao and Guo, Sitong and Xie, Zhongwei and Bao, Hujun and Xu, Weiwei and Zou, Changqing},
  booktitle = ICCV,
  year = {2025}
}

@article{lu2025see4d,
  title = {{See4D: Pose-Free 4D Generation via Auto-Regressive Video Inpainting}},
  author = {Lu, Dongyue and Liang, Ao and Huang, Tianxin and Fu, Xiao and Zhao, Yuyang and Ma, Baorui and Pan, Liang and Yin, Wei and Kong, Lingdong and Ooi, Wei Tsang and Liu, Ziwei},
  journal = CGF,
  year = {2026}
}

@inproceedings{lin2026vista4d,
  title = {{Vista4D: Video Reshooting with 4D Point Clouds}},
  author = {Lin, Kuan Heng and Liu, Zhizheng and Salamanca, Pablo and Kant, Yash and Burgert, Ryan and Xu, Yuancheng and Namekata, Koichi and Zhao, Yiwei and Zhou, Bolei and Goldblum, Micah and Debevec, Paul and Yu, Ning},
  booktitle = CVPR,
  year = {2026}
}

@inproceedings{liu2026mocam,
  title = {{MoCam: Unified Novel View Synthesis via Structured Denoising Dynamics}},
  author = {Liu, Haofeng and Zhou, Yang and Wang, Ziheng and Xu, Zhengbo and Peng, Zhan and Ma, Jie and Liang, Jun and He, Shengfeng and Li, Jing},
  booktitle = ECCV,
  year = {2026}
}

@inproceedings{chen2026depthdirector,
  title = {{Beyond Inpainting: Unleash 3D Understanding for Stable Camera-Controlled Video Re-Rendering}},
  author = {Chen, Dong-Yu and Guo, Yixin and Yang, Shuojin and Mu, Tai-Jiang and Hu, Shi-Min},
  booktitle = ECCV,
  year = {2026}
}

@inproceedings{cai2026scaling4d,
  title = {{Scaling4D: Pushing the Frontier of Video Novel View Synthesis through Large-Scale Monocular Videos}},
  author = {Cai, Hongrui and Luo, Junjie and Fu, Zhihong and Zhu, Shengnan and Wen, Jiawei and Feng, Wanquan and Zhao, Songtao and He, Qian},
  booktitle = CVPR,
  year = {2026}
}

@inproceedings{dit,
  title = {{Scalable Diffusion Models with Transformers}},
  author = {Peebles, William and Xie, Saining},
  booktitle = ICCV,
  year = {2023}
}

@inproceedings{zhang2024monst3r,
  title = {{MonST3R: A Simple Approach for Estimating Geometry in the Presence of Motion}},
  author = {Zhang, Junyi and Herrmann, Charles and Hur, Junhwa and Jampani, Varun and Darrell, Trevor and Cole, Forrester and Sun, Deqing and Yang, Ming-Hsuan},
  booktitle = ICLR,
  year = {2025}
}

@inproceedings{zhang2023adding,
  title = {{Adding Conditional Control to Text-to-Image Diffusion Models}},
  author = {Zhang, Lvmin and Rao, Anyi and Agrawala, Maneesh},
  booktitle = ICCV,
  year = {2023}
}

@inproceedings{corona2021meva,
  title = {{MEVA: A Large-Scale Multiview, Multimodal Video Dataset for Activity Detection}},
  author = {Corona, Kellie and Osterdahl, Katie and Collins, Roderic and Hoogs, Anthony},
  booktitle = WACV,
  year = {2021}
}

@inproceedings{grauman2022ego4d,
  title = {{Ego4D: Around the World in 3,000 Hours of Egocentric Video}},
  author = {Grauman, Kristen and Westbury, Andrew and Byrne, Eugene and Chavis, Zachary and Furnari, Antonino and Girdhar, Rohit and Hamburger, Jackson and Jiang, Hao and Liu, Miao and Liu, Xingyu and Martin, Miguel and Nagarajan, Tushar and Radosavovic, Ilija and Ramakrishnan, Santhosh Kumar and Ryan, Fiona and Sharma, Jayant and Wray, Michael and Xu, Mengmeng and Xu, Eric Zhongcong and Zhao, Chen and Bansal, Siddhant and Batra, Dhruv and Cartillier, Vincent and Crane, Sean and Do, Tien and Doulaty, Morrie and Erapalli, Akshay and Feichtenhofer, Christoph and Fragomeni, Adriano and Fu, Qichen and Gebreselasie, Abrham and Gonz\'alez, Cristina and Hillis, James and Huang, Xuhua and Huang, Yifei and Jia, Wenqi and Khoo, Weslie and Kol\'a\v{r}, J\'achym and Kottur, Satwik and Kumar, Anurag and Landini, Federico and Li, Chao and Li, Yanghao and Li, Zhenqiang and Mangalam, Karttikeya and Modhugu, Raghava and Munro, Jonathan and Murrell, Tullie and Nishiyasu, Takumi and Price, Will and Ruiz, Paola and Ramazanova, Merey and Sari, Leda and Somasundaram, Kiran and Southerland, Audrey and Sugano, Yusuke and Tao, Ruijie and Vo, Minh and Wang, Yuchen and Wu, Xindi and Yagi, Takuma and Zhao, Ziwei and Zhu, Yunyi and Arbel\'aez, Pablo and Crandall, David and Damen, Dima and Farinella, Giovanni Maria and Fuegen, Christian and Ghanem, Bernard and Ithapu, Vamsi Krishna and Jawahar, C. V. and Joo, Hanbyul and Kitani, Kris and Li, Haizhou and Newcombe, Richard and Oliva, Aude and Park, Hyun Soo and Rehg, James M. and Sato, Yoichi and Shi, Jianbo and Shou, Mike Zheng and Torralba, Antonio and Torresani, Lorenzo and Yan, Mingfei and Malik, Jitendra},
  booktitle = CVPR,
  year = {2022}
}

@inproceedings{zheng2023pointodyssey,
  title = {{PointOdyssey: A Large-Scale Synthetic Dataset for Long-Term Point Tracking}},
  author = {Zheng, Yang and Harley, Adam W. and Shen, Bokui and Wetzstein, Gordon and Guibas, Leonidas J.},
  booktitle = ICCV,
  year = {2023}
}

@inproceedings{sener2022assembly101,
  title = {{Assembly101: A Large-Scale Multi-View Video Dataset for Understanding Procedural Activities}},
  author = {Sener, Fadime and Chatterjee, Dibyadip and Shelepov, Daniel and He, Kun and Singhania, Dipika and Wang, Robert and Yao, Angela},
  booktitle = CVPR,
  year = {2022}
}

@inproceedings{greff2022kubric,
  title = {{Kubric: A Scalable Dataset Generator}},
  author = {Greff, Klaus and Belletti, Francois and Beyer, Lucas and Doersch, Carl and Du, Yilun and Duckworth, Daniel and Fleet, David J. and Gnanapragasam, Dan and Golemo, Florian and Herrmann, Charles and Kipf, Thomas and Kundu, Abhijit and Lagun, Dmitry and Laradji, Issam and Liu, Hsueh-Ti (Derek) and Meyer, Henning and Miao, Yishu and Nowrouzezahrai, Derek and Oztireli, Cengiz and Pot, Etienne and Radwan, Noha and Rebain, Daniel and Sabour, Sara and Sajjadi, Mehdi S. M. and Sela, Matan and Sitzmann, Vincent and Stone, Austin and Sun, Deqing and Vora, Suhani and Wang, Ziyu and Wu, Tianhao and Yi, Kwang Moo and Zhong, Fangcheng and Tagliasacchi, Andrea},
  booktitle = CVPR,
  year = {2022}
}

@inproceedings{nan2024openvid,
  title = {{OpenVid-1M: A Large-Scale High-Quality Dataset for Text-to-video Generation}},
  author = {Nan, Kepan and Xie, Rui and Zhou, Penghao and Fan, Tiehan and Yang, Zhenheng and Chen, Zhijie and Li, Xiang and Yang, Jian and Tai, Ying},
  booktitle = ICLR,
  year = {2025}
}

@inproceedings{vda,
  title = {{Video Depth Anything: Consistent Depth Estimation for Super-Long Videos}},
  author = {Chen, Sili and Guo, Hengkai and Zhu, Shengnan and Zhang, Feihu and Huang, Zilong and Feng, Jiashi and Kang, Bingyi},
  booktitle = CVPR,
  year = {2025}
}

@inproceedings{ling2024dl3dv,
  title = {{DL3DV-10K: A Large-Scale Scene Dataset for Deep Learning-based 3D Vision}},
  author = {Ling, Lu and Sheng, Yichen and Tu, Zhi and Zhao, Wentian and Xin, Cheng and Wan, Kun and Yu, Lantao and Guo, Qianyu and Yu, Zixun and Lu, Yawen and Li, Xuanmao and Sun, Xingpeng and Ashok, Rohan and Mukherjee, Aniruddha and Kang, Hao and Kong, Xiangrui and Hua, Gang and Zhang, Tianyi and Benes, Bedrich and Bera, Aniket},
  booktitle = CVPR,
  year = {2024}
}

@article{zhou2018real10k,
  title = {{Stereo Magnification: Learning View Synthesis Using Multiplane Images}},
  author = {Zhou, Tinghui and Tucker, Richard and Flynn, John and Fyffe, Graham and Snavely, Noah},
  journal = TOG,
  year = {2018}
}

@misc{CarveKit,
  title = {{CarveKit}},
  author = {{OPHoperHPO}},
  howpublished = {\url{https://github.com/OPHoperHPO/image-background-remove-tool}},
  year = {n.d.}
}

@inproceedings{sam2,
  title = {{SAM 2: Segment Anything in Images and Videos}},
  author = {Ravi, Nikhila and Gabeur, Valentin and Hu, Yuan-Ting and Hu, Ronghang and Ryali, Chaitanya and Ma, Tengyu and Khedr, Haitham and R\"{a}dle, Roman and Rolland, Chloe and Gustafson, Laura and Mintun, Eric and Pan, Junting and Alwala, Kalyan Vasudev and Carion, Nicolas and Wu, Chao-Yuan and Girshick, Ross and Dollar, Piotr and Feichtenhofer, Christoph},
  booktitle = ICLR,
  year = {2025}
}

@article{miao2025rose,
  title = {{ROSE: Remove Objects with Side Effects in Videos}},
  author = {Miao, Chenxuan and Feng, Yutong and Zeng, Jianshu and Gao, Zixiang and Liu, Hantang and Yan, Yunfeng and Qi, Donglian and Chen, Xi and Wang, Bin and Zhao, Hengshuang},
  journal = {arXiv:2508.18633},
  year = {2025}
}

@inproceedings{reizenstein21co3d,
  title = {{Common Objects in 3D: Large-Scale Learning and Evaluation of Real-Life 3D Category Reconstruction}},
  author = {Reizenstein, Jeremy and Shapovalov, Roman and Henzler, Philipp and Sbordone, Luca and Labatut, Patrick and Novotny, David},
  booktitle = ICCV,
  year = {2021}
}

@inproceedings{zhang2018unreasonable,
  title = {{The Unreasonable Effectiveness of Deep Features as a Perceptual Metric}},
  author = {Zhang, Richard and Isola, Phillip and Efros, Alexei A. and Shechtman, Eli and Wang, Oliver},
  booktitle = CVPR,
  year = {2018}
}

@inproceedings{iphone,
  title = {{Monocular Dynamic View Synthesis: A Reality Check}},
  author = {Gao, Hang and Li, Ruilong and Tulsiani, Shubham and Russell, Bryan and Kanazawa, Angjoo},
  booktitle = NIPS,
  year = {2022}
}

@article{hu2025ex4d,
  title = {{EX-4D: EXtreme Viewpoint 4D Video Synthesis via Depth Watertight Mesh}},
  author = {Hu, Tao and Peng, Haoyang and Liu, Xiao and Ma, Yuewen},
  journal = {arXiv:2506.05554},
  year = {2025}
}

@inproceedings{song2025worldforge,
  title = {{Taming Video Models for 3D and 4D Generation via Zero-Shot Camera Control}},
  author = {Song, Chenxi and Yang, Yanming and Zhao, Tong and Li, Ruibo and Zhang, Chi},
  booktitle = CVPR,
  year = {2026}
}

@inproceedings{freeorbit4d,
  title = {{FreeOrbit4D: Training-Free Arbitrary Camera Redirection for Monocular Videos via Foreground-Complete 4D Reconstruction}},
  author = {Cao, Wei and Zhang, Hao and Tian, Fengrui and Wu, Yulun and Li, Yingying and Wang, Shenlong and Yu, Ning and Liu, Yaoyao},
  booktitle = SIGGRAPH,
  year = {2026}
}

@inproceedings{reshootanything,
  title = {{Reshoot-Anything: A Self-Supervised Model for In-the-Wild Video Reshooting}},
  author = {Paliwal, Avinash and Iyer, Adithya and Yadav, Shivin and Afridi, Muhammad Ali and Harikumar, Midhun},
  booktitle = CVPRW,
  year = {2026}
}

@inproceedings{vbench,
  title = {{VBench: Comprehensive Benchmark Suite for Video Generative Models}},
  author = {Huang, Ziqi and He, Yinan and Yu, Jiashuo and Zhang, Fan and Si, Chenyang and Jiang, Yuming and Zhang, Yuanhan and Wu, Tianxing and Jin, Qingyang and Chanpaisit, Nattapol and Wang, Yaohui and Chen, Xinyuan and Wang, Limin and Lin, Dahua and Qiao, Yu and Liu, Ziwei},
  booktitle = CVPR,
  year = {2024}
}

@inproceedings{dreamscene4d,
  title = {{DreamScene4D: Dynamic Multi-Object Scene Generation from Monocular Videos}},
  author = {Chu, Wen-Hsuan and Ke, Lei and Fragkiadaki, Katerina},
  booktitle = NIPS,
  year = {2024}
}

@misc{worldscapemoe2026,
  title = {{WorldScape-0.2 (MoE) (WorldScore leaderboard submission)}},
  author = {{WorldScape Team}},
  howpublished = {\url{https://worldscape.io/}},
  year = {2026}
}

@misc{worlddreamerentry2026,
  title = {{World Dreamer (WorldScore leaderboard submission)}},
  author = {{World Dreamer Team}},
  howpublished = {\url{https://github.com/worldAgents-c}},
  year = {2026}
}

@misc{evophysworld2026,
  title = {{EvoPhys-World}},
  author = {{EvoPhys Team}},
  howpublished = {\url{https://evophys.com/}},
  year = {2026}
}

@article{inspatioworld2026,
  title = {{INSPATIO-WORLD: A Real-Time 4D World Simulator via Spatiotemporal Autoregressive Modeling}},
  author = {{InSpatio Team} and Shen, Donghui and Zhang, Guofeng and Liu, Haomin and Ji, Haoyu and Bao, Hujun and Zhai, Hongjia and Liu, Jialin and Guo, Jing and Wang, Nan and Pan, Siji and Pan, Weihong and Xie, Weijian and Liu, Xianbin and Xiang, Xiaojun and Zhang, Xiaoyu and Chen, Xinyu and Wang, Yifu and Chen, Yipeng and Fan, Zhenzhou and Le, Zhewen and Ye, Zhichao and Zhao, Ziqiang},
  journal = {arXiv:2604.07209},
  year = {2026}
}

@misc{eonworldentry2026,
  title = {{EonWorld (WorldScore leaderboard submission)}},
  author = {{EonWorld Team}},
  howpublished = {\url{https://github.com/X-EraAI}},
  year = {2026}
}

@inproceedings{dai2026fantasyworld,
  title = {{FantasyWorld: Geometry-Consistent World Modeling via Unified Video and 3D Prediction}},
  author = {Dai, Yixiang and Jiang, Fan and Wang, Chiyu and Xu, Mu and Qi, Yonggang},
  booktitle = ICLR,
  year = {2026}
}

@article{chen2025teleworld,
  title = {{TeleWorld: Towards Dynamic Multimodal Synthesis with a 4D World Model}},
  author = {Chen, Yabo and Liang, Yuanzhi and Wang, Jiepeng and Chen, Tingxi and Cheng, Junfei and Gu, Zixiao and Huang, Yuyang and Jiang, Zicheng and Li, Wei and Li, Tian and Li, Weichen and Li, Zuoxin and Liu, Guangce and Liu, Jialun and Liu, Junqi and Wang, Haoyuan and Weng, Qizhen and Wu, Xuan'er and Xiang, Xunzhi and Yang, Xiaoyan and Zhang, Xin and Zhang, Shiwen and Zhou, Junyu and Zhou, Chengcheng and Huang, Haibin and Zhang, Chi and Li, Xuelong},
  journal = {arXiv:2601.00051},
  year = {2025}
}

@article{huang2025voyager,
  title = {{Voyager: Long-Range and World-Consistent Video Diffusion for Explorable 3D Scene Generation}},
  author = {Huang, Tianyu and Zheng, Wangguandong and Wang, Tengfei and Liu, Yuhao and Wang, Zhenwei and Wu, Junta and Jiang, Jie and Li, Hui and Lau, Rynson and Zuo, Wangmeng and Guo, Chunchao},
  journal = TOG,
  year = {2025}
}
